\documentclass{article} 
\usepackage{iclr2026_conference,times}

\usepackage{hyperref}
\usepackage{url}

\usepackage[utf8]{inputenc}
\usepackage[T1]{fontenc}
\usepackage{hyperref}
\usepackage{url}
\usepackage{booktabs}
\usepackage{amsfonts}
\usepackage{amsmath}
\usepackage{amssymb}
\usepackage{nicefrac}
\usepackage{microtype}
\usepackage{xcolor}
\usepackage{graphicx}
\usepackage{subcaption}
\usepackage{multirow}
\usepackage{caption}
\usepackage{enumitem}
\usepackage{xspace}
\usepackage{wrapfig}

\newcommand{\modelname}{\textsf{ContextEA}\xspace}

\title{Harnessing Structural Context for Entity Alignment Foundation Models}

\author{
  Xingyu Chen$^{1}$, Yuanning Cui$^{2}$, Zequn Sun$^{1}$\thanks{Corresponding author}, Wei Hu$^{1,3}$\\
  $^{1}$State Key Laboratory for Novel Software Technology, Nanjing University, Nanjing, China\\
  $^{2}$Nanjing University of Information Science and Technology, Nanjing, China\\
  $^{3}$National Institute of Healthcare Data Science, Nanjing University, Nanjing, China \\
  \texttt{xingyuchen@smail.nju.edu.cn, yncui@nuist.edu.cn} \\
  \texttt{\{sunzq, whu\}@nju.edu.cn} \\
}

\iclrfinalcopy 

\begin{document}

\maketitle

\begin{abstract}
Entity alignment (EA) aims to identify equivalent entities across heterogeneous knowledge graphs (KGs) and is a key component of knowledge fusion and cross-KG reasoning. The recent EA foundation model demonstrates that alignment knowledge, once pretrained, can be directly applied to diverse previously unseen KG pairs. However, it still underuses structural context in two places: cross-KG interaction is weak during encoding, and final candidate ranking still relies too heavily on coarse similarity. We address these limitations with \modelname, an enhanced encoder-decoder framework for transferable EA. On the encoder side, we introduce a cross-KG interaction encoder that unifies the two KGs with anchor bridges and performs earlier relation-aware cross-graph propagation. On the decoder side, we introduce a structural calibration decoder that calibrates alignment scores with entity-level, neighborhood-level, relation-level, and anchor-aware structural evidence. This design strengthens both structural context construction and structural context exploitation while remaining lightweight. Experiments on 29 EA datasets in OpenEA, SRPRS, and DBP show consistent gains over strong transferable baselines. Notably, the pretrained \modelname already surpasses the finetuned baselines on all three benchmark groups, demonstrating substantially stronger transfer to unseen KGs. These results suggest that explicitly harnessing structural context is an effective direction for improving EA foundation models.
\end{abstract}

\section{Introduction}
Knowledge graphs (KGs) built from different sources often describe overlapping entities with different identifiers, schemas, and local structures \citep{KG_survey}. 
Entity alignment (EA) aims to identify entities across such KGs that refer to the same real-world object \citep{MTransE,BootEA,NoisyEA24,Lambda,NeuSymEA}, and therefore serves as a basic component of knowledge fusion, cross-KG retrieval, KG completion, and downstream KG applications~\citep{EA_survey,chen2025language}. 
Classical EA methods are usually trained for a fixed KG pair with some seed alignments as anchors \citep{EA_embed_survey}.
In contrast, practical applications increasingly require \emph{transferable} EA: a model should learn alignment knowledge from source KG pairs and then generalize to unseen target KGs with different scales, densities, and heterogeneity patterns.

This demand has recently motivated the development of EA foundation models. 
EAFM~\citep{EAFM} is a recent representative model, as it shifts EA from KG-specific fitting toward a pretrain-and-transfer reasoning paradigm. 
Technically, EAFM first learns relation-aware structural representations, then encodes the two KGs with anchor-conditioned propagation in each KG, and finally ranks cross-KG candidates with a lightweight matching head.
While this direction is promising, it also raises a more challenging question: what types of structural context should be transferred across unseen KGs, and how should such context be effectively utilized?
We argue that the current bottleneck is not merely model capacity. Instead, the primary limitation lies in the underutilization of structural context, both during representation learning and in the final decision-making process.

\begin{wrapfigure}{r}{0.6\textwidth}
\centering
\includegraphics[width=0.9999\linewidth]{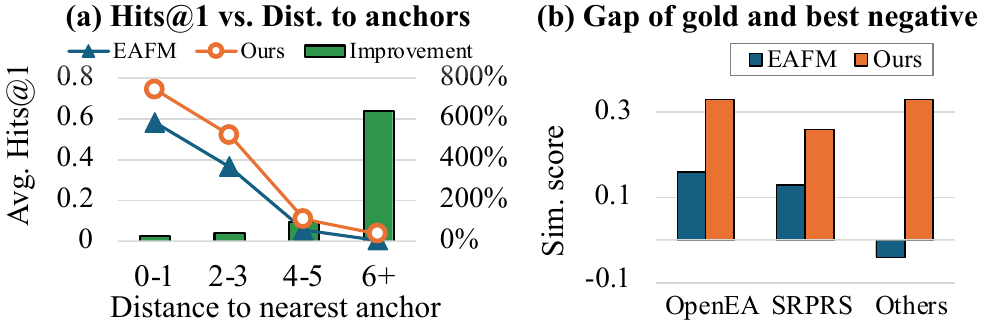}
\caption{(a) EA performance drops substantially as the query entity becomes farther from the training anchors, indicating that transferable EA is highly sensitive to the availability of cross-KG structural context. 
(b) In EAFM, the score gap between the gold target and the hardest negative remains very small or even negative. Our structural calibration can enlarge the final score gap and improves discrimination.}
\label{fig:intro}
\end{wrapfigure}

The first limitation lies in \emph{delayed and weak cross-KG interaction during encoding}. 
In EAFM, each KG is mostly processed separately, and interactions between them are delayed until the final matching stage. 
As a result, entity representations are shaped primarily by intra-KG information, while useful signals from the other KG arrive too late to contribute effectively. 
This limitation becomes critical in the transferable setting: generalization to unseen KGs requires entities to capture and integrate cross-KG relational information during representation learning. 
Figure~\ref{fig:intro} (a) illustrates the issue, showing a clear drop in performance as query entities move further from training anchors.

The second limitation concerns \emph{structural disambiguation during decoding}. 
Even when the gold entity has already been retrieved into a small top-$k$ candidate set, similarity-based ranking may still remain structurally ambiguous. 
In practice, several candidates may receive very similar coarse scores, while only one is truly consistent with the query under local neighborhood patterns, relation semantics, and anchor-supported correspondences. 
Figure~\ref{fig:intro} (b) shows this issue: 
the coarse gap between the gold target and the hardest negative remains very small, and sometimes even negative.
This suggests that a transferable EA model should not only construct better structural context during encoding, but also exploit it more explicitly during decoding.

Motivated by these observations, we propose \modelname, a unified framework for \emph{harnessing structural context} in transferable EA. 
Our method contains two tightly coupled modules. 
On the encoder side, we introduce a \emph{cross-KG interaction encoder} that unifies the two KGs through anchor bridges and performs relation-aware cross-graph propagation during representation learning. 
Instead of delaying interaction until the final matching stage, this encoder allows structural evidence to flow across KGs earlier, so that entity representations are learned under richer structure context. 
On the decoder side, we introduce a \emph{structural calibration decoder} that explicitly verifies whether a top-ranked candidate is structurally plausible. 
It integrates entity-level, neighborhood-level, relation-level, and anchor-aware evidence, and outputs a structural correction score that calibrates the original coarse score.

A key point of \modelname is that the encoder and decoder play complementary roles. 
The encoder brings in cross-KG context earlier through anchor bridges, and the decoder uses this context to recheck top candidates.
In this sense, the encoder helps build better structural representations, while the decoder helps make the final decision more reliable.
\modelname is also lightweight. 
Instead of heavy cross-attention or full subgraph matching, we use pooled neighborhood, relation, and anchor-aware summaries for alignment score calibration. This keeps EA easy to interpret and analyze. 

The contributions of this work are summarized as follows:
\begin{itemize}
    \item We revisit transferable EA from the perspective of structural context and identify two coupled bottlenecks in existing work: 
    weak cross-KG interaction during entity encoding 
    and insufficient structural disambiguation during alignment decoding.
    
    \item We propose \modelname, a lightweight encoder-decoder framework. 
    Its cross-KG interaction encoder improves cross-KG structural propagation during representation learning. 
    Its structural calibration decoder uses the resulting structural evidence for EA verification.
    
    \item We achieve strong results on 29 datasets from OpenEA \citep{OpenEA}, SRPRS \citep{rsn4ea}, and DBP \citep{JAPE,BootEA}. 
    The pretrained \modelname already outperforms the finetuned EAFM baseline. We further provide analyses on graph characteristics, efficiency, ablations, and parameter sensitivity to show why the method works.
\end{itemize}

\section{Related Work}

\paragraph{Knowledge graph foundation models.}
Recent KG reasoning research has shifted from transductive embedding models toward transferable graph models based on relative structural patterns rather than dataset-specific entity embeddings. Earlier inductive methods reduce reliance on fixed entity identities by reasoning over relational paths and local subgraphs \citep{NBFNet,REDGNN,Conditional_Message_Passing}. Related studies also explore transferable representation learning through multi-source pretraining and local adaptation \citep{multisouce_pretraining}.
Building on this trend, recent KG foundation models target broader cross-graph generalization. ULTRA learns transferable reasoning through a relation-centric design and supports zero-shot transfer to unseen KGs \citep{ULTRA}. KG-ICL further studies prompt-based in-context reasoning on KGs \citep{KG-ICL}. Recent studies also enhance cross-graph transferability through more expressive transferable representations \citep{MOTIF,TRIX}. EAFM is the closest prior to our setting, as it adapts the KGFM paradigm to EA through anchor-conditioned parallel encoding and a learnable interaction matcher \citep{EAFM}. Our work follows the transferable spirit of EAFM, but focuses on how to better inject and exploit structural context throughout both encoding and decoding.

\paragraph{Entity alignment.}
EA has evolved from translation-based embedding methods to structure-aware graph matching models. 
Early studies such as MTransE learn cross-KG correspondences in a shared embedding space \citep{MTransE}. Later methods strengthen structural modeling with graph neural networks and relation-aware interaction \citep{GCNAlign,AliNet,rrea,Dual-AMN,SelfKG,RePS}. 
IMEA further improve local context modeling or transferability \citep{IMEA}. 
Recent studies also revisit embedding-based EA from the perspective of robustness and adaptivity \citep{LIME,ClusterEA,RobustAdaptiveEA}, and analyze entity similarity under multi-source KG settings \citep{SimilarityFlooding}. 
In recent years, some work exploits large language models for EA \citep{NoisyEA24,LLM-Align,EasyEA}.
Despite their strong performance, most existing EA methods are still optimized for a specific KG pair, and their final decisions remain largely dominated by coarse similarity. 
EAFM takes an important step toward transferable EA by enabling zero-shot generalization across unseen KGs \citep{EAFM}.
This line of work motivates our focus on how to inject and exploit structural context more effectively.

\section{Preliminaries}
\label{sec:preliminaries}

\paragraph{Entity alignment.}
Let $\mathcal{G}_1=(\mathcal{E}_1,\mathcal{R}_1,\mathcal{T}_1)$ and $\mathcal{G}_2=(\mathcal{E}_2,\mathcal{R}_2,\mathcal{T}_2)$ denote two KGs, where $\mathcal{E}$, $\mathcal{R}$ and $\mathcal{T}$ denote the sets of entities, relations and facts, respectively.
Let $\mathcal{S}=\{(e_i^{(1)}\in\mathcal{E}_1,e_i^{(2)}\in\mathcal{E}_2)\}$ be a set of one-to-one seed alignments between $\mathcal{G}_1$ and $\mathcal{G}_2$.
EA aims to identify, for each query entity $q \in \mathcal{E}_1$, its counterpart in $\mathcal{E}_2$. In practice, the EA model learns a scoring function $s(q, c)$ over candidates $c \in \mathcal{E}_2$, and the final prediction is obtained by ranking candidates according to this score.

\paragraph{Transferable entity alignment.}
Transferable EA considers a set of source tasks $\mathcal{D}_{\mathrm{src}}=\{(\mathcal{G}_1^\mathrm{s},\mathcal{G}_2^\mathrm{s},\mathcal{S}^\mathrm{s})\}$ and requires the model to generalize to unseen target KG pairs. 
The target pair still provides a small seed set to facilitate direct transfer.
The key requirement is that the model should rely on transferable structural cues rather than text-based entity identities, 
so that it can remain effective when the entity vocabulary, relation vocabulary, graph scale, and graph heterogeneity all change.

\paragraph{Entity alignment foundation model.}
EAFM formulates transferable EA as a query-conditioned, relation-guided cross-KG reasoning problem. Its key idea is to avoid memorizing KG-specific entity embeddings and instead represent an unseen KG through relative structural patterns.
Given a query entity $q$ and two KGs $\mathcal{G}_1, \mathcal{G}_2$,
EAFM first encodes relations in a query-conditioned manner. Specifically, the relation encoder is initialized with a query-aware signal: relations connected to the query entity $q$ are initialized as all-one vectors, while all other relations are initialized as all-zero vectors, thereby explicitly injecting query-specific conditions into the subsequent relation propagation process.
It builds a merged relation graph over $\mathcal{R}_1 \cup \mathcal{R}_2$ and computes relation representations
\begin{equation}
\mathbf{R}(q) = f_{\mathrm{rel}}(\mathcal{G}_r, q),
\end{equation}
where $\mathcal{G}_r$ is the relation graph that captures relation-to-relation interactions induced by aligned anchors,
$\mathbf{R}(q) \in \mathbb{R}^{|\mathcal{R}_1 \cup \mathcal{R}_2| \times d}$ denotes relation representation matrix conditioning on entity $q$.
In this way, the learned relation representations capture patterns relevant to the current alignment task and remain applicable to unseen entities and local structures in the target KG.

Next, EAFM performs query-conditioned entity propagation on the two KGs. Seed alignments $\mathcal{S}$ are converted into cross-KG bridge edges, and the entity encoder is initialized with a query-aware entity prior: anchor entities within the 2-hop neighborhood of the query entity $q$, together with their aligned counterparts connected through $\mathcal{S}$, are initialized as all-one vectors, while all remaining entities are initialized as all-zero vectors. This initialization explicitly injects query-centered structural context into the entity propagation process and guides information flow toward query-relevant regions across the two KGs. The entity encoder then computes
\begin{equation}
\mathbf{H}_1(q), \mathbf{H}_2(q) = f_{\mathrm{ent}}(\mathcal{G}_1, \mathcal{G}_2, \mathcal{S}, \mathbf{R}(q), q).
\end{equation}
Here, each node update depends not only on its neighbors but also on the current query $q$, so messages are adaptively weighted and combined according to the query context. This ensures that the same propagation rule can be applied to unseen KGs without learning new entity-specific embeddings.

Finally, a lightweight matcher scores each query-candidate pair from the encoded entity states. This query-conditioned, relation-guided encoder-plus-matcher design is the technical foundation of our method and explains how knowledge learned from one KG can be transferred to another unseen KG.

\paragraph{Query-conditioned message passing.}
At the core of EAFM is query-conditioned message passing. Each node $i$ updates its state based on neighbors $\mathcal{N}(i)$ and the current query entity $q$:
\begin{equation}
    \mathbf{h}_i^{(\ell+1)}
    =
    \psi\Big(
        \mathbf{h}_i^{(\ell)},
        \sum_{j \in \mathcal{N}(i)} \alpha_{ji}^{(\ell)}
        \phi(\mathbf{h}_j^{(\ell)}, \mathbf{e}_{ji}, q)
    \Big),
\end{equation}
where $q$ conditions both message construction $\phi$ and neighbor weighting $\alpha$, allowing the propagation to focus on paths and structures relevant to the current query. By stacking multiple layers of such query-conditioned updates, the encoder can interpret unseen KGs in the context of the query rather than relying on fixed entity IDs. This is the mechanism that enables EAFM-style transferability.

\section{Method}

\subsection{Framework Overview}
\begin{figure}[t]
\centering
\includegraphics[width=0.999\textwidth]{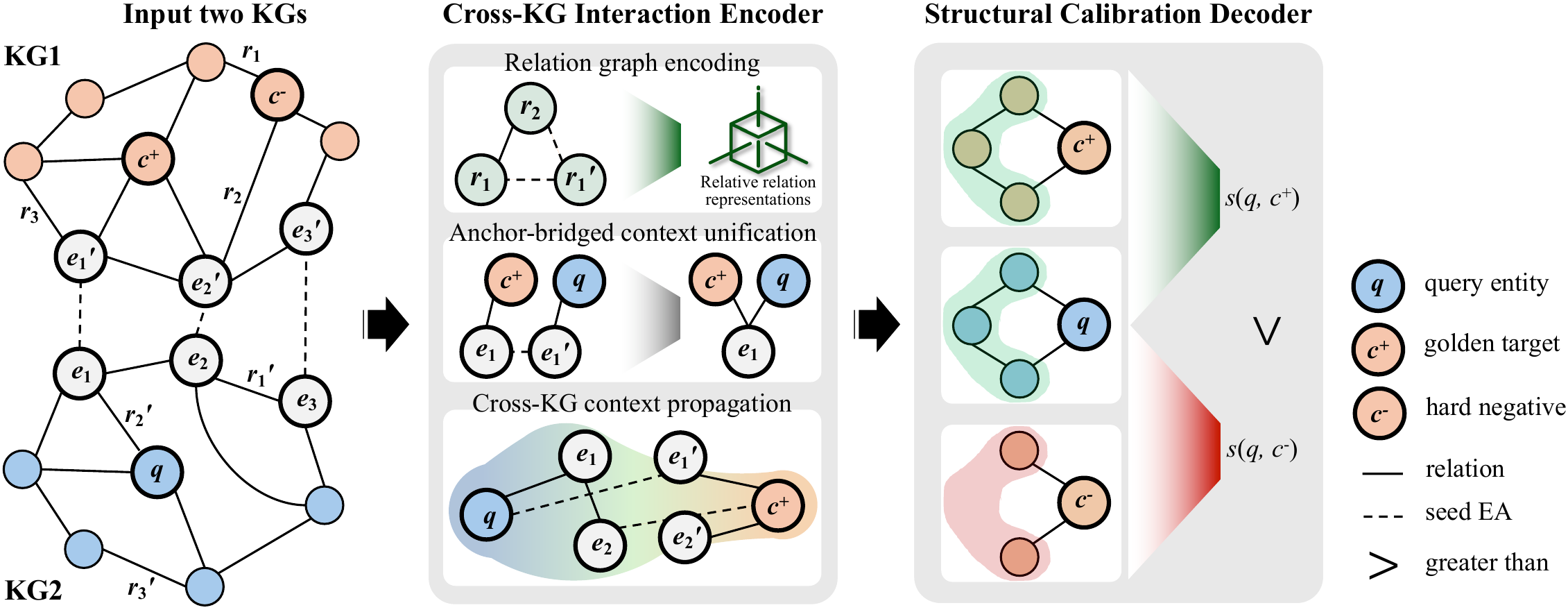}
\caption{
Framework overview of \modelname. 
It first builds cross-KG structural representations and then calibrates top candidates with structural evidence.
} 
\label{fig:framework}
\end{figure}

Given two KGs and a set of training anchors,
our method follows a lightweight encoder-decoder pipeline for harnessing structural context. 
As illustrated in Figure~\ref{fig:framework},
the cross-KG interaction encoder first learns relation-aware states, instantiates anchor bridges, and performs cross-KG propagation on the unified graph to obtain enriched entity representations. 
A lightweight coarse matcher then retrieves a top-$k$ candidate set for each query, 
and the structural calibration decoder further verifies these candidates with entity-level, neighborhood-level, relation-level, and anchor-aware structural evidence.
The final score is obtained by adding a structural correction term to the coarse score.

\subsection{Cross-KG Interaction Encoder}
In EAFM,
the two KGs are encoded mostly in isolation and cross-KG interaction is delayed to the final matching stage. 
We instead reformulate the encoding stage as a \emph{cross-KG interaction encoder}, so that cross-KG structural evidence can enter representation learning earlier.

\paragraph{Anchor-bridged relation graph encoding.}
We first construct a relation graph, where nodes denote relations and edges represent structural co‑occurrence patterns between relations, such as shared anchor entities and shared head or tail entities.
We then encode relation representations using a stack of query-conditioned relation graph neural layers, 
yielding 
$\mathbf{R}(q) \in \mathbb{R}^{|\mathcal{R}_1 \cup \mathcal{R}_2| \times d}$.  
These representations provide relation- and query-aware structural context for subsequent entity propagation.

\paragraph{Anchor-bridged graph unification.}
We then unify the two KGs using the training alignments. 
Specifically, each anchor EA pair $(e_i^{(1)}, e_i^{(2)}) \in \mathcal{S}$ is instantiated as an explicit cross-KG bridge edge. 
Original KG edges preserve intra-KG structure, while anchor bridges enable structural context to move across graphs during encoding. 
This step transforms the two separate KGs into a unified graph.

\paragraph{Cross-KG interaction propagation.}
On top of the unified graph, we perform query- and relation-conditioned message passing over entities. Because anchor bridges are already present during encoding, each entity can absorb not only intra-KG neighborhood evidence, but also cross-KG structural evidence mediated by aligned anchors. Abstractly, the encoder can be written as
\begin{equation}
    \mathbf{H}(q) = f_{\mathrm{enc}}\big(\mathcal{G}_1,\mathcal{G}_2,\mathcal{S},\mathbf{R}(q),q\big),
\end{equation}
where $\mathbf{H}(q) \in \mathbb{R}^{|\mathcal{E}_1 \cup \mathcal{E}_2| \times d}$ denotes the final query-conditioned entity representations. 
Compared with EAFM, this encoder strengthens cross-KG interaction at the representation-learning stage and supplies richer structural context to the downstream decoder.

\subsection{Structural Calibration Decoder}

Our decoder exploits structural context for final pairwise verification. 
We first perform coarse retrieval with the base matcher. 
For a query-candidate pair $(q,c)$, the coarse matching score is computed as
\begin{equation}
    s_{\mathrm{emb}}(q,c)
    =
    [\lvert \mathbf{h}_q - \mathbf{h}_c \rvert ; \mathbf{h}_c]
    \mathbf{W}_{\mathrm{emb}},
\end{equation}
where $\mathbf{h}_q, \mathbf{h}_c \in \mathbb{R}^d$ are the entity embeddings from the encoder, $[\cdot;\cdot]$ denotes concatenation, $\lvert \mathbf{h}_q - \mathbf{h}_c \rvert$ is the element-wise absolute difference, and $\mathbf{W}_{\mathrm{emb}} \in \mathbb{R}^{2d \times 1}$ is a learnable linear scoring matrix. The resulting scalar score $s_{\mathrm{emb}}(q,c) \in \mathbb{R}$ is used to retrieve the top-$k$ candidates for each query.

However, coarse similarity alone often remains insufficient to distinguish the gold target from structurally confusing hard negatives. 
We therefore calibrate the similarities with a structural decoder.

\textbf{Direct entity compatibility.}
The first structural view is direct entity compatibility, encoded by
\begin{equation}
    \phi_{\mathrm{ent}}(q,c) = [\lvert \mathbf{h}_q - \mathbf{h}_c \rvert ; \mathbf{h}_q \odot \mathbf{h}_c],
\end{equation}
where $\odot$ denotes element-wise product and captures whether the query and candidate are semantically consistent at the entity level.

\textbf{Neighborhood structure.}
The second view summarizes local neighborhood structure from neighborhood semantics. We mean-pool one-hop entity representations:
\begin{equation}
    \mathbf{g}_q = \mathrm{MeanPool}\big(\{\mathbf{h}_n \mid n \in \mathcal{N}(q)\}\big), \qquad
    \mathbf{g}_c = \mathrm{MeanPool}\big(\{\mathbf{h}_n \mid n \in \mathcal{N}(c)\}\big),
\end{equation}
where $\mathcal{N}(q), \mathcal{N}(c)$ are the sets of one-hop neighbor entities of $q$ and $c$. We then define
\begin{equation}
    \phi_{\mathrm{nei}}(q,c) = [\lvert \mathbf{g}_q - \mathbf{g}_c \rvert ; \mathbf{g}_q \odot \mathbf{g}_c].
\end{equation}

\textbf{Relation patterns.}
The third view summarizes local structure from relation patterns. We mean-pool one-hop relation representations:
\begin{equation}
    \mathbf{r}_q = \mathrm{MeanPool}\big(\{\mathbf{r}_\ell \mid \ell \in \mathcal{R}(q)\}\big), \qquad
    \mathbf{r}_c = \mathrm{MeanPool}\big(\{\mathbf{r}_\ell \mid \ell \in \mathcal{R}(c)\}\big),
\end{equation}
where $\mathcal{R}(q), \mathcal{R}(c)$ are the sets of one-hop relations adjacent to $q$ and $c$, and $\mathbf{r}_\ell \in \mathbb{R}^d$ is the relation embedding. We define
\begin{equation}
    \phi_{\mathrm{rel}}(q,c) = [\lvert \mathbf{r}_q - \mathbf{r}_c \rvert ; \mathbf{r}_q \odot \mathbf{r}_c],
\end{equation}
to capture whether the candidate is supported by compatible local relation context.

\textbf{Anchor-supported local structure.}
The fourth view explicitly models anchor-supported local structure. We identify one-hop neighbor pairs backed by training anchors, denote the supported subsets by $\mathcal{A}_q$ and $\mathcal{A}_c$, and compute pooled anchor-aware summaries
\begin{equation}
    \mathbf{z}_{aq} = \operatorname{MeanPool}(\mathcal{A}_q), \qquad
    \mathbf{z}_{ac} = \operatorname{MeanPool}(\mathcal{A}_c),
\end{equation}
as well as a set of scalar anchor statistics, denoted by $\phi_{anc}^{scalar}(q,c)$, including supported-neighbor ratios, support strength, relation consistency, and supported-entity similarity. The resulting anchor-aware feature is
\begin{equation}
    \phi_{anc}(q,c)
    =
    [\mathbf{z}_{aq};
     \mathbf{z}_{ac};
     \lvert \mathbf{z}_{aq} - \mathbf{z}_{ac} \rvert;
     \mathbf{z}_{aq} \odot \mathbf{z}_{ac};
     \phi_{anc}^{scalar}(q,c)].
\end{equation}
This view captures how strongly the candidate pair is supported by anchor-induced structural context.

\paragraph{Structural feature combination.}
We concatenate four structure views into a unified representation
\begin{equation}
    \phi(q,c) = [\phi_{\mathrm{ent}}(q,c); \phi_{\mathrm{nei}}(q,c); \phi_{\mathrm{rel}}(q,c); \phi_{\mathrm{anc}}(q,c)].
\end{equation}
To fuse these heterogeneous structural cues, we employ a lightweight multi-layer perceptron, which outputs the correction score
$
\Delta(q,c) = f_{\mathrm{dec}}\big(\phi(q,c)\big).
$
This acts as a compact structural aggregation head that transforms multiple views of structural context into a single correction score.

\paragraph{Final score calibration.}
The final calibrated score is obtained by adding the structural correction to the original coarse score:
\begin{equation}
    s_{\mathrm{final}}(q,c) = s_{\mathrm{emb}}(q,c) + \beta \cdot \Delta(q,c),
\end{equation}
where $\beta \in \mathbb{R}$ controls the strength of structural correction. In this way, the decoder does not replace the original matching head, but explicitly calibrates it with richer structural context, making it especially useful for resolving hard negatives that remain ambiguous under coarse similarity.

\subsection{Training}
To optimize transferable EA, we train the model using a bidirectional contrastive objective over the source alignment set $\mathcal{S}^\mathrm{s}$. 
For each aligned entity pair $(e_i, e_j) \in \mathcal{S}^\mathrm{s}$, the final matching score $s_{\mathrm{final}}(\cdot,\cdot)$ is optimized by maximizing the probability of retrieving the correct counterpart entity from the opposite KG:
\begin{equation}
\small
\begin{aligned}
    \mathcal{L} = - \frac{1}{|\mathcal{S}^\mathrm{s}|}
    \sum_{(e_i, e_j) \in \mathcal{S}^\mathrm{s}}
    \Bigg[
    \log 
    \frac{\exp(s_{\mathrm{final}}(e_i, e_j))}
    {\sum_{e_k \in \mathcal{E}_2} \exp(s_{\mathrm{final}}(e_i, e_k))}
    +\;
    \log 
    \frac{\exp(s_{\mathrm{final}}(e_j, e_i))}
    {\sum_{e_k \in \mathcal{E}_1} \exp(s_{\mathrm{final}}(e_j, e_k))}
    \Bigg],
\end{aligned}
\end{equation}
where $\mathcal{E}_1$ and $\mathcal{E}_2$ denote the entity sets of the two KGs, respectively. 
This objective jointly enforces correct alignment retrieval in both directions and encourages the learned representations to capture transferable structural correspondence rather than KG-specific entity identities.

Overall, the proposed framework follows a coherent principle for harnessing structural context: 
the encoder enriches structural context through stronger cross-KG interaction, and the decoder further exploits this enriched structural context to improve final candidate discrimination.

\section{Experiments}

\subsection{Experimental Settings}

\paragraph{Datasets.}
Following EAFM \citep{EAFM},
we evaluate on 29 EA datasets from three benchmark groups: 
OpenEA \citep{OpenEA}, SRPRS \citep{rsn4ea}, and DBpedia-based benchmarks (DBP) \citep{MTransE,BootEA}. 
Appendix ~\ref{appx:statistics} summarizes the statistics of these datasets.
They cover cross-KG and cross-lingual alignment, graph scales from 15K to 100K, and both sparse (V1) and dense (V2) structural regimes. Following EAFM, we use an inductive protocol: the model is pretrained on source KG pairs and then transferred to unseen target KG pairs either directly (``pretrain'') or with target adaptation (``finetune''). 
The aligned entity pairs in each dataset are split into 20\%, 10\%, and 70\% for training, validation, and testing, respectively.

\paragraph{Implementations.}
Following EAFM, we use D-W-15K-V1 and EN-DE-15K-V1 as source datasets so that pretraining covers both cross-KG and cross-lingual structural patterns. 
We adopt the same training configuration as EAFM for fair comparison, including a 6-layer relation encoder, a 6-layer entity encoder, hidden size 32, 
balance coefficient $\beta=0.3$, AdamW optimizer with learning rate $5\times 10^{-4}$, batch size 64, and early stopping based on validation MRR.
We set the number of candidates $k=16$ in our decoder for a efficiency-effectiveness trade-off.
The code is in supplementary material.

\paragraph{Metrics and implementations.}
We report Mean Reciprocal Rank (MRR), Hits@10 and Hits@1 results to assess EA performance. 
We treat Hits@1 as the most direct measure of exact EA results. 
The main baselines are ULTRA-EA (a EA variant of ULTRA \citep{ULTRA}) and EAFM \citep{EAFM}, both evaluated under the same ``pretrain'' and ``finetune'' settings.

\subsection{Main Results}

\begin{table*}[t]
\centering
\caption{Comparison of average EA results across different dataset groups, where the \textbf{bold} and \underline{underline} indicate the best and second-best results, respectively.}
\resizebox{\textwidth}{!}{
\begin{tabular}{lcccccccccccc}
\toprule
\multirow{2}{*}{{Method}}
& \multicolumn{3}{c}{{OpenEA}}
& \multicolumn{3}{c}{{SRPRS}}
& \multicolumn{3}{c}{{DBP}}
& \multicolumn{3}{c}{{Average}} \\
\cmidrule(lr){2-4} \cmidrule(lr){5-7} \cmidrule(lr){8-10} \cmidrule(lr){11-13}
& MRR & H@10 & H@1
& MRR & H@10 & H@1
& MRR & H@10 & H@1
& MRR & H@10 & H@1 \\
\midrule
ULTRA-EA pretrain
& 0.477 & 0.635 & 0.385
& 0.350 & 0.557 & 0.241
& 0.203 & 0.379 & 0.107
& 0.395 & 0.569 & 0.297 \\
ULTRA-EA finetune
& 0.509 & 0.691 & 0.416
& 0.380 & 0.587 & 0.271
& 0.223 & 0.409 & 0.137
& 0.424 & 0.614 & 0.328\\
\midrule
EAFM \ pretrain
& 0.602 & 0.764 & 0.517
& 0.530 & 0.735 & 0.426
& 0.294 & 0.452 & 0.214
& 0.529 & 0.702 & 0.440 \\
EAFM \ finetune
& 0.656 & 0.814 & 0.573
& 0.605 & 0.792 & 0.508
& 0.464 & 0.662 & 0.394
& 0.609 & 0.782 & 0.524\\
\midrule
\modelname \ pretrain
& \underline{0.713} & \underline{0.857} & \underline{0.636}
& \underline{0.675} & \underline{0.837} & \underline{0.587}
& \underline{0.522} & \underline{0.667} & \underline{0.444}
& \underline{0.669} & \underline{0.819} & \underline{0.589} \\
\modelname \ finetune
& \textbf{0.722} & \textbf{0.859} & \textbf{0.648}
& \textbf{0.680} & \textbf{0.839} & \textbf{0.594}
& \textbf{0.588} & \textbf{0.743} & \textbf{0.503}
& \textbf{0.688} & \textbf{0.834} & \textbf{0.608} \\
\bottomrule
\end{tabular}
}
\label{tab:group_results}
\end{table*}

Table~\ref{tab:group_results} summarizes the average results of each EA groups. 
The full detailed results of each dataset are in Appendix~\ref{appx:results}.
Three general patterns can be observed. 
(i) \modelname consistently outperforms the baselines across all dataset groups under both pretraining and finetuning settings. 
In the pretraining scenario, it shows clear improvements over EAFM. 
Notably, the pretrained \modelname already exceeds the performance of the finetuned EAFM baseline on all benchmark groups.
This suggests that the proposed method possesses strong transferability even before adaptation to target KGs.
(ii) Finetuning further enhances performance. 
The gains are relatively moderate on OpenEA and SRPRS, whereas a more pronounced improvement is observed on DBP.
This indicates that target-specific adaptation plays a greater role in more challenging settings.
(iii) A clear performance gap remains between \modelname and ULTRA-EA under both settings. 
This result suggests that EA-oriented pretraining alone is insufficient when the model architecture is not explicitly tailored to entity alignment, highlighting the need to model cross-KG correspondences.

\begin{table*}[h]
\centering
\caption{MRR results across different graph characteristics. The datasets are categorized by density (Sparse vs. Dense), scale (15K vs. 100K), and heterogeneity type (Cross-KG vs. Cross-Lingual).
}
\label{tab:detailed_breakdown}
\resizebox{0.85\textwidth}{!}{
\begin{tabular}{lcccccc}
\toprule
\multirow{2}{*}{{Method}}
& \multicolumn{2}{c}{{Graph Density}}
& \multicolumn{2}{c}{{Graph Scale}}
& \multicolumn{2}{c}{{Heterogeneity}} \\
\cmidrule(lr){2-3} \cmidrule(lr){4-5} \cmidrule(lr){6-7}
& {V1 (Sparse)} & {V2 (Dense)}
& {15K} & {100K}
& {Cross-KG} & {Cross-Lingual} \\
\midrule
ULTRA-EA pretrain
& 0.314 & 0.509
& 0.393 & 0.397
& 0.350 & 0.203 \\
ULTRAEA finetune
& 0.342 & 0.540
& 0.420 & 0.431
& 0.530 & 0.325 \\
\midrule
EAFM \ pretrain
& 0.451 & 0.639
& 0.532 & 0.523
& 0.597 & 0.465 \\
EAFM \ finetune
& 0.509 & 0.692
& 0.602 & 0.552
& 0.649 & 0.524 \\
\midrule
\modelname \ pretrain
& 0.598 & 0.806
& 0.726 & 0.674
& 0.737 & 0.623 \\
\modelname \ finetune
& 0.600 & 0.812
& 0.724 & 0.693
& 0.744 & 0.628 \\
\bottomrule
\end{tabular}
}
\end{table*}

\subsection{Detailed Analysis on Graph Characteristics}

Table~\ref{tab:detailed_breakdown} analyzes performance under different graph characteristics. 
\modelname consistently performs best across all categories.
(i) On sparse KGs, the pretrained model already outperforms the finetuned EAFM baseline, showing that stronger cross‑KG interaction helps when structural cues are weak.
On dense KGs, the advantage becomes even larger, suggesting that the encoder–decoder design makes better use of richer structure contexts.
(ii) A similar trend is observed across graph scales. \modelname performs well on both small and large datasets, with larger gains on bigger graphs.
This indicates that cross‑KG propagation and structural calibration become more important as the candidate space grows.
(iii) Our method also remains stable under different heterogeneity levels.
These results show that \modelname generalizes well across variations in density, scale, and heterogeneity.

\subsection{Ablation Study}

\paragraph{Effectiveness of our encoder and decoder.}
Table~\ref{tab:ablation_results} evaluates the roles of our encoder and decoder.
Replacing the decoder with the coarse matcher leads to a moderate drop in performance.
This shows that although the coarse matcher is strong, the decoder still adds useful structural calibration for distinguishing hard negatives, especially on challenging datasets such as DBP.
In contrast, replacing our encoder with the EAFM encoder causes a much larger decline across all metrics.
The drop is clearly greater than that caused by removing the decoder, again with the strongest impact on DBP.
This indicates that cross‑KG interaction during encoding is the main source of transfer ability, while the decoder mainly improves the final ranking.
Overall, the encoder and decoder play complementary roles: the encoder drives transferability, and the decoder sharpens EA discrimination.

\begin{table*}[t]
\centering
\caption{Ablation results across dataset groups. In the variant ``w/o Our encoder'', we use the EAFM encoder instead. In the variant ``w/o Our decoder'', we use coarse similarities for decoding.}
\resizebox{\textwidth}{!}{\large
\begin{tabular}{lcccccccccccc}
\toprule
\multirow{2}{*}{{Method}}
& \multicolumn{3}{c}{{OpenEA}}
& \multicolumn{3}{c}{{SRPRS}}
& \multicolumn{3}{c}{{DBP}}
& \multicolumn{3}{c}{{Average}} \\
\cmidrule(lr){2-4} \cmidrule(lr){5-7} \cmidrule(lr){8-10} \cmidrule(lr){11-13}
& MRR & H@10 & H@1
& MRR & H@10 & H@1
& MRR & H@10 & H@1
& MRR & H@10 & H@1 \\
\midrule
\modelname\
& {0.713} & {0.857} & {0.636}
& {0.675} & {0.837} & {0.587}
& {0.522} & {0.667} & {0.444}
& {0.669} & {0.819} & {0.589} \\
\ \ w/o Our decoder
& {0.696} & {0.843} & {0.618}
& {0.666} & {0.834} & {0.577}
& {0.493} & {0.643} & {0.412}
& {0.653} & {0.806} & {0.571} \\
\ \ w/o Our encoder
& 0.565 & 0.731 & 0.479
& 0.535 & 0.741 & 0.429
& 0.240 & 0.361 & 0.177
& 0.501 & 0.670 & 0.413 \\
\bottomrule
\end{tabular}
}
\label{tab:ablation_results}
\end{table*}

\begin{wrapfigure}{r}{0.35\textwidth}
\centering
\vspace{-10pt}
\includegraphics[width=0.98\linewidth]{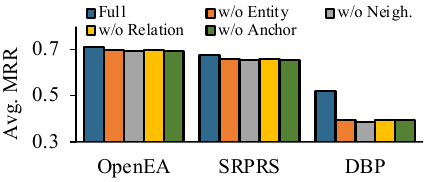}
\caption{Ablation study on structure views used in our decoder.}
\label{fig:views}
\vspace{-10pt}
\end{wrapfigure}

\paragraph{Effectiveness of different structure contexts in our decoder.}
Figure~\ref{fig:views} further studies the four structure views used in our decoder. 
Detailed results are in Appendix~\ref{appx:structureviews}.
The full model performs best on all groups. 
Removing any single view causes a drop, which shows that each view provides useful structural evidence. The impact is relatively small on OpenEA and SRPRS, where the full model is already strong, but becomes much clearer on DBP.
This suggests that difficult datasets benefit more from explicit structural calibration.

\begin{figure*}[h]
\centering
\begin{subfigure}[t]{0.245\textwidth}
\centering
\includegraphics[width=\linewidth]{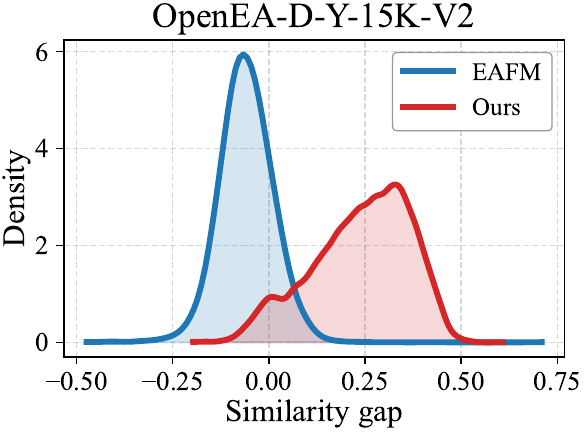}
\end{subfigure}
\begin{subfigure}[t]{0.245\textwidth}
\centering
\includegraphics[width=\linewidth]{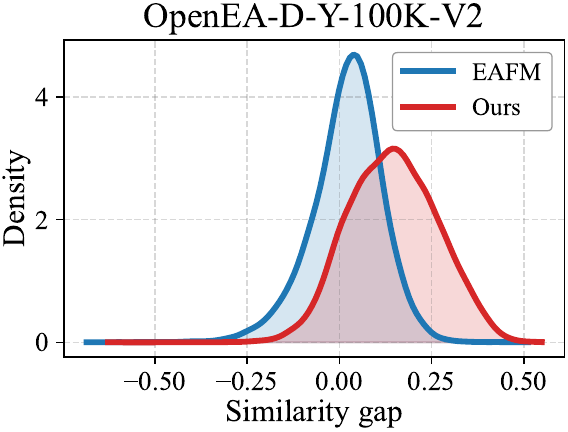}
\end{subfigure}
\begin{subfigure}[t]{0.245\textwidth}
\centering
\includegraphics[width=\linewidth]{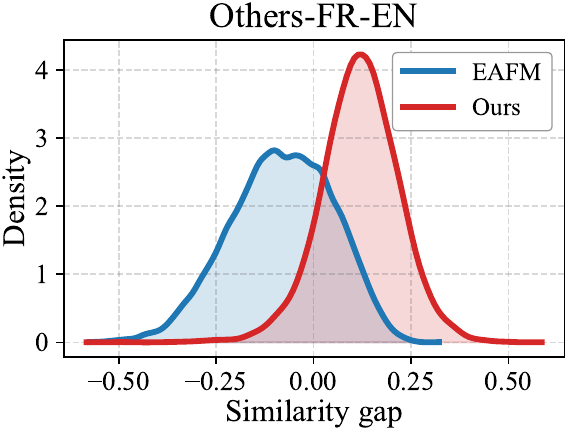}
\end{subfigure}
\begin{subfigure}[t]{0.245\textwidth}
\centering
\includegraphics[width=\linewidth]{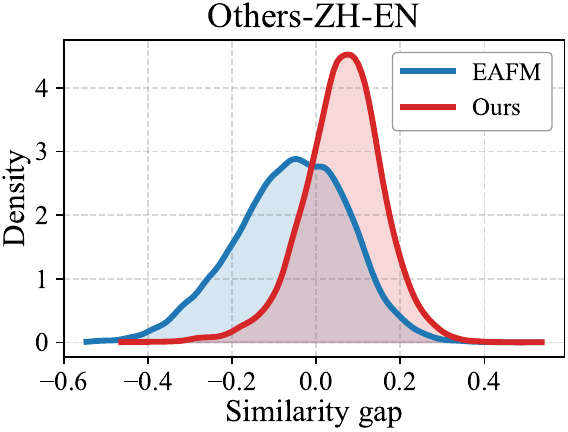}
\end{subfigure}
\caption{Density plots of the similarity gap between the gold target and the hardest negative candidate on four datasets. 
Results on other datasets are in Appendix \ref{sec:appendix_gap_distributions}.
}
\label{fig:gap_distribution}
\end{figure*}

\subsection{Other Analyses}
\paragraph{Gold vs. hard-negative score gaps.}
Figure~\ref{fig:gap_distribution} shows how \modelname changes pairwise discrimination.
For each query entity, we compute the similarity gap between the gold target and the hardest negative candidate. 
Larger positive values indicate clearer separation, while negative values mean that the hardest negative is ranked above the true target.
We can see that the density curve of \modelname is shifted to the right, meaning that more queries obtain a larger positive margin.
These results complement the ranking metrics in Tables~\ref{tab:group_results}--\ref{tab:ablation_results}.
The improvement is not only in average MRR or Hits@1, but also in the score distribution: the model creates a clearer gap between the gold entity and the hardest negative.
This matches the purpose of the structural calibration decoder, which enlarges the score gap when coarse similarity is not enough.

\paragraph{Efficiency analysis.}

\begin{wrapfigure}{r}{0.45\textwidth}
\centering
\includegraphics[width=0.95\linewidth]{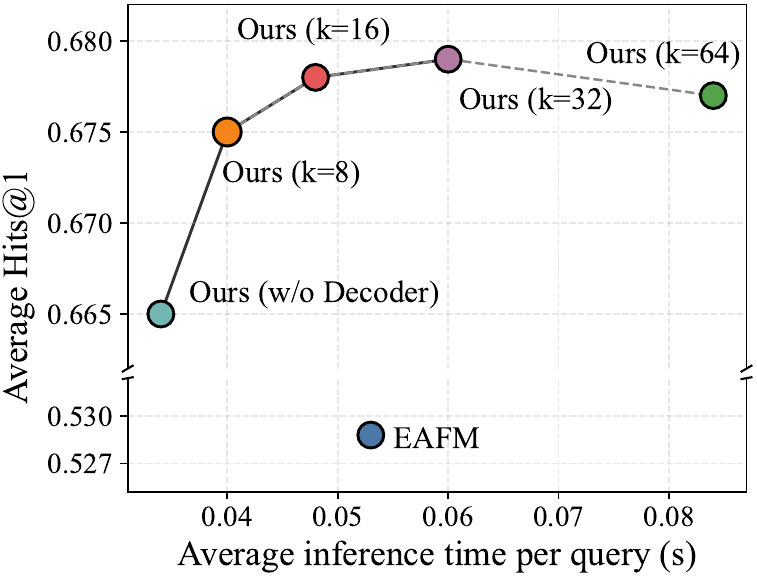}
\caption{Efficiency-effectiveness trade-off.}
\label{fig:efficiency_tradeoff}
\end{wrapfigure}

Figure~\ref{fig:efficiency_tradeoff} compares the efficiency–effectiveness trade-off of \modelname. 
Detailed results are presented in Appendix \ref{appx:efficiency}.
The experiments are conducted on an NVIDIA RTX 4090 GPU.
Efficiency is measured as the average inference time per query. 
We vary the size $k$ to examine its impact.
The fastest variant is ``Ours (w/o Decoder)'', which also achieves a large gain over EAFM in Hits@1 while reducing inference time. 
This shows that our encoder is not only more accurate, but also more efficient than the original EAFM inference pipeline. 
Adding the decoder increases inference time, but it further improves accuracy over the decoder-free variant. 
The best trade-off is achieved at small to medium candidate sizes, while larger $k$ mainly adds cost with limited benefit. 
Overall, the results support our claim that structural calibration is lightweight and effective when applied to a small candidate set.

\paragraph{Sensitivity analysis.}

\begin{wrapfigure}{r}{0.35\textwidth}
\centering
\vspace{-12pt}
\includegraphics[width=0.9\linewidth]{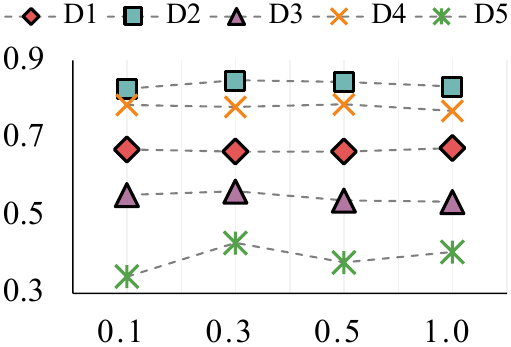}
\caption{MRR w.r.t. weight $\beta$.}
\label{fig:beta}
\vspace{-12pt}
\end{wrapfigure}

Figure~\ref{fig:beta} studies the sensitivity of the structural calibration weight $\beta$. 
Detailed results are in Appendix ~\ref{appx:beta}.
The overall trend is stable, which indicates that the decoder does not rely on brittle hyperparameter tuning. 
A moderate value around $\beta=0.3$ gives the best average performance, suggesting that structural calibration should act as a correction to the coarse score rather than dominate it. 
When $\beta$ is too small, structural evidence is underused; when it becomes too large, some datasets show mild over-correction. 
This is consistent with our goal of keeping the decoder lightweight but effective.

\paragraph{Case study.}

\begin{figure*}[h]
\centering
\includegraphics[width=\textwidth]{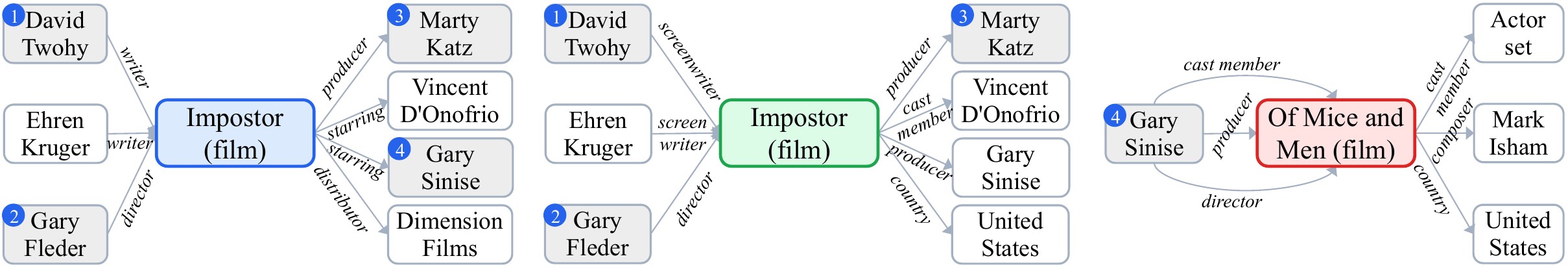}
\caption{A case study of EA. The left subgraph shows the query entity. 
The middle green node is the correct matched entity in the target KG, and the right red node is a hard negative candidate. Blue numbered nodes indicate anchor entities.}
\label{fig:case_study}
\end{figure*}

Figure~\ref{fig:case_study} shows a typical successful alignment made by \modelname. 
The query entity is correctly aligned to the green target entity, while the red entity is a hard negative.
The hard negative is heavily influenced by the anchor ``Gary Sinise'', which provides three relations around it. 
The anchor also has a direct link to the query. 
This gives the wrong candidate strong local support.
EAFM cannot clearly separate the two candidates using coarse similarity.
In contrast, our encoder passes local semantic and relational signals to build a clearer context.
The decoder then benefits from this improved structural context.
Thus, \modelname can produce a clear score gap.

\section{Conclusion and Future Work}
In this paper, we study transferable EA from the perspective of structural context. 
We identify two limitations in existing work: weak cross-KG interaction during encoding and limited structural discrimination during final ranking. 
To address them, we propose \modelname, a lightweight encoder-decoder framework with a cross-KG interaction encoder and a structural calibration decoder.
Experiments on OpenEA, SRPRS, and DBP show clear improvements. 
There are several directions for future work. One is to design stronger propagation modules for larger and more heterogeneous KGs. Another is to build richer anchor-aware structures for final calibration. 

\section*{Acknowledgments}
This work was supported by the National Natural Science Foundation of China (Nos. 62406136 and 62272219) and the Natural Science Foundation of Jiangsu Province (No. BK20241246).

\bibliographystyle{iclr2026_conference}
\bibliography{custom}

\newpage
\appendix

\section{Statistics of EA datasets}\label{appx:statistics}
We list the statistics of all EA datasets in Table~\ref{tab:ea_datasets}.
It includes three benchmark collections: OpenEA, SRPRS, and DBP.
For each dataset, we show the two paired KGs and report the numbers of entities, relations, and triples.
These benchmarks cover different KG pairs, languages, and graph sizes, from 15K to 100K entities, with varying structural densities.
This diversity supports a systematic evaluation under different alignment settings.

\begin{table*}[h]
\centering
\caption{Statistics of EA datasets.}
\label{tab:ea_datasets}
\resizebox{0.9999\textwidth}{!}{
\begin{tabular}{llcrrrcrrr}
\toprule
\multirow{2}{*}{\textbf{Group}} & \multirow{2}{*}{\textbf{Dataset}} &
\multicolumn{4}{c}{$\mathcal{G}_1$} &
\multicolumn{4}{c}{$\mathcal{G}_2$} \\
\cmidrule(lr){3-6}\cmidrule(lr){7-10}
 &  & KG & \# Ent & \# Rel & \# Facts & KG & \# Ent & \# Rel & \# Facts \\
\midrule

\multirow{16}{*}{OpenEA}
& EN-DE-15K-V1   & DBpedia (EN) & 15,000  & 215 & 47,676  & DBpedia (DE) & 15,000  & 131 & 50,419  \\
& EN-DE-15K-V2   & DBpedia (EN) & 15,000  & 169 & 84,867  & DBpedia (DE) & 15,000  &  96 & 92,632  \\
& EN-FR-15K-V1   & DBpedia (EN) & 15,000  & 267 & 47,334  & DBpedia (FR) & 15,000  & 210 & 40,864  \\
& EN-FR-15K-V2   & DBpedia (EN) & 15,000  & 193 & 96,318  & DBpedia (FR) & 15,000  & 166 & 80,112  \\
& D-W-15K-V1     & DBpedia      & 15,000  & 248 & 38,265  & Wikidata     & 15,000  & 169 & 42,746  \\
& D-W-15K-V2     & DBpedia      & 15,000  & 167 & 73,983  & Wikidata     & 15,000  & 121 & 83,365  \\
& D-Y-15K-V1     & DBpedia      & 15,000  & 165 & 30,291  & YAGO3        & 15,000  &  28 & 26,638  \\
& D-Y-15K-V2     & DBpedia      & 15,000  &  72 & 68,063  & YAGO3        & 15,000  &  21 & 60,970  \\
& EN-DE-100K-V1  & DBpedia (EN) & 100,000 & 381 & 335,359 & DBpedia (DE) & 100,000 & 196 & 336,240 \\
& EN-DE-100K-V2  & DBpedia (EN) & 100,000 & 323 & 622,588 & DBpedia (DE) & 100,000 & 170 & 629,395 \\
& EN-FR-100K-V1  & DBpedia (EN) & 100,000 & 400 & 309,607 & DBpedia (FR) & 100,000 & 300 & 258,285 \\
& EN-FR-100K-V2  & DBpedia (EN) & 100,000 & 379 & 649,902 & DBpedia (FR) & 100,000 & 287 & 561,391 \\
& D-W-100K-V1    & DBpedia      & 100,000 & 413 & 293,990 & Wikidata     & 100,000 & 261 & 251,708 \\
& D-W-100K-V2    & DBpedia      & 100,000 & 318 & 616,457 & Wikidata     & 100,000 & 239 & 588,203 \\
& D-Y-100K-V1    & DBpedia      & 100,000 & 287 & 294,188 & YAGO3        & 100,000 &  32 & 400,518 \\
& D-Y-100K-V2    & DBpedia      & 100,000 & 230 & 576,547 & YAGO3        & 100,000 &  31 & 865,265 \\
\midrule

\multirow{8}{*}{SRPRS}
& D-W-V1   & DBpedia (EN) & 15,000 & 253 & 38,421 & Wikidata (EN) & 15,000 & 144 & 40,159 \\
& D-W-V2   & DBpedia (EN) & 15,000 & 220 & 68,598 & Wikidata (EN) & 15,000 & 135 & 75,465 \\
& D-Y-V1   & DBpedia (EN) & 15,000 & 219 & 33,571 & YAGO3 (EN)    & 15,000 &  30 & 34,660 \\
& D-Y-V2   & DBpedia (EN) & 15,000 & 206 & 71,257 & YAGO3 (EN)    & 15,000 &  30 & 97,131 \\
& EN-DE-V1 & DBpedia (EN) & 15,000 & 225 & 38,281 & DBpedia (DE)  & 15,000 & 118 & 37,069 \\
& EN-DE-V2 & DBpedia (EN) & 15,000 & 207 & 56,983 & DBpedia (DE)  & 15,000 & 117 & 59,848 \\
& EN-FR-V1 & DBpedia (EN) & 15,000 & 221 & 36,508 & DBpedia (FR)  & 15,000 & 177 & 33,532 \\
& EN-FR-V2 & DBpedia (EN) & 15,000 & 217 & 71,929 & DBpedia (FR)  & 15,000 & 174 & 66,760 \\
\midrule

\multirow{5}{*}{DBP}
& ZH-EN & DBpedia (ZH) & 19,388 & 1,701 & 70,414  & DBpedia (EN) & 19,572 & 1,323 & 95,142  \\
& JA-EN & DBpedia (JA) & 19,814 & 1,299 & 77,214  & DBpedia (EN) & 19,780 & 1,153 & 93,484  \\
& FR-EN & DBpedia (FR) & 19,661 &   903 & 105,998 & DBpedia (EN) & 19,993 & 1,208 & 115,722 \\
& D-W-100K
  & DBpedia   & 100,000 & 330 & 463,294
  & Wikipedia & 100,000 & 220 & 448,774 \\
& D-Y-100K
  & DBpedia & 100,000 & 302 & 428,952
  & YAGO3   & 100,000 &  31 & 502,563 \\
\bottomrule
\end{tabular}
}
\end{table*}

\section{Detailed EA Results}
\label{appx:results}

We present the detailed results for each EA dataset in Table~\ref{tab:detailed_ea_results}.
The table reports MRR and Hits@10 (H@10) for all benchmark groups, including OpenEA, SRPRS, and DBpedia.
All results follow the same evaluation protocol described in the main text.

\begin{table*}[t]
\centering
\caption{Detailed experimental results on EA benchmarks.}
\label{tab:detailed_ea_results}
\resizebox{\textwidth}{!}{\Large
\begin{tabular}{llcccccccccccccccc}
\toprule
\multirow{2}{*}{\textbf{Group}} & \multirow{2}{*}{\textbf{Dataset}}
& \multicolumn{2}{c}{\textbf{ULTRA-EA}}
& \multicolumn{2}{c}{\textbf{ULTRA-EA}}
& \multicolumn{2}{c}{\textbf{EAFM}}
& \multicolumn{2}{c}{\textbf{EAFM}}
& \multicolumn{3}{c}{\textbf{\modelname}}
& \multicolumn{3}{c}{\textbf{\modelname}} \\
& 
& \multicolumn{2}{c}{\textbf{(pretrain)}}
& \multicolumn{2}{c}{\textbf{(finetune)}}
& \multicolumn{2}{c}{\textbf{(pretrain)}}
& \multicolumn{2}{c}{\textbf{(finetune)}}
& \multicolumn{3}{c}{\textbf{(pretrain)}}
& \multicolumn{3}{c}{\textbf{(finetune)}} \\
\cmidrule(lr){3-4}\cmidrule(lr){5-6}\cmidrule(lr){7-8}\cmidrule(lr){9-10}\cmidrule(lr){11-13}\cmidrule(lr){14-16}
& & MRR & H@10 & MRR & H@10 & MRR & H@10 & MRR & H@10 & MRR & H@10 & H@1 & MRR & H@10 & H@1 \\
\midrule
\multirow{16}{*}{OpenEA}
& D-W-15K-V1   & 0.431 & 0.613 & 0.461 & 0.713 & 0.604 & 0.795 & 0.582 & 0.777 & 0.674 & 0.847 & 0.578 & 0.665 & 0.839 & 0.570 \\
& D-W-15K-V2   & 0.671 & 0.852 & 0.701 & 0.902 & 0.738 & 0.890 & 0.753 & 0.908 & 0.842 & 0.961 & 0.774 & 0.840 & 0.959 & 0.773 \\
& D-Y-15K-V1   & 0.560 & 0.699 & 0.590 & 0.749 & 0.636 & 0.770 & 0.664 & 0.787 & 0.704 & 0.809 & 0.636 & 0.699 & 0.802 & 0.631 \\
& D-Y-15K-V2   & 0.901 & 0.936 & 0.951 & 0.986 & 0.950 & 0.987 & 0.959 & 0.990 & 0.971 & 0.994 & 0.958 & 0.972 & 0.994 & 0.958 \\
& D-W-100K-V1  & 0.342 & 0.539 & 0.372 & 0.619 & 0.459 & 0.650 & 0.514 & 0.711 & 0.548 & 0.733 & 0.455 & 0.564 & 0.747 & 0.471 \\
& D-W-100K-V2  & 0.568 & 0.749 & 0.598 & 0.809 & 0.437 & 0.590 & 0.630 & 0.791 & 0.694 & 0.853 & 0.608 & 0.732 & 0.880 & 0.651 \\
& D-Y-100K-V1  & 0.558 & 0.731 & 0.588 & 0.791 & 0.694 & 0.853 & 0.730 & 0.873 & 0.740 & 0.879 & 0.665 & 0.745 & 0.878 & 0.673 \\
& D-Y-100K-V2  & 0.772 & 0.849 & 0.802 & 0.919 & 0.762 & 0.888 & 0.876 & 0.946 & 0.908 & 0.971 & 0.871 & 0.916 & 0.973 & 0.883 \\
& EN-DE-15K-V1 & 0.587 & 0.780 & 0.617 & 0.830 & 0.737 & 0.900 & 0.729 & 0.890 & 0.786 & 0.934 & 0.710 & 0.782 & 0.931 & 0.705 \\
& EN-DE-15K-V2 & 0.754 & 0.862 & 0.784 & 0.912 & 0.765 & 0.889 & 0.805 & 0.914 & 0.886 & 0.969 & 0.835 & 0.887 & 0.967 & 0.838 \\
& EN-FR-15K-V1 & 0.254 & 0.487 & 0.284 & 0.547 & 0.533 & 0.782 & 0.528 & 0.778 & 0.620 & 0.847 & 0.504 & 0.617 & 0.840 & 0.502 \\
& EN-FR-15K-V2 & 0.267 & 0.477 & 0.297 & 0.537 & 0.460 & 0.686 & 0.572 & 0.814 & 0.723 & 0.924 & 0.613 & 0.702 & 0.907 & 0.592 \\
& EN-DE-100K-V1 & 0.386 & 0.532 & 0.416 & 0.592 & 0.520 & 0.682 & 0.557 & 0.728 & 0.573 & 0.744 & 0.489 & 0.583 & 0.745 & 0.502 \\
& EN-DE-100K-V2 & 0.248 & 0.434 & 0.278 & 0.464 & 0.594 & 0.736 & 0.674 & 0.796 & 0.746 & 0.863 & 0.683 & 0.757 & 0.867 & 0.697 \\
& EN-FR-100K-V1 & 0.178 & 0.330 & 0.208 & 0.360 & 0.364 & 0.572 & 0.431 & 0.647 & 0.439 & 0.630 & 0.347 & 0.445 & 0.637 & 0.351 \\
& EN-FR-100K-V2 & 0.163 & 0.298 & 0.193 & 0.328 & 0.378 & 0.552 & 0.491 & 0.684 & 0.588 & 0.760 & 0.495 & 0.628 & 0.802 & 0.536 \\
\midrule
\multirow{8}{*}{SRPRS}
& D-W-V1    & 0.258 & 0.432 & 0.288 & 0.462 & 0.340 & 0.559 & 0.431 & 0.647 & 0.514 & 0.724 & 0.410 & 0.522 & 0.731 & 0.418 \\
& D-W-V2    & 0.489 & 0.755 & 0.519 & 0.785 & 0.633 & 0.827 & 0.719 & 0.885 & 0.819 & 0.951 & 0.742 & 0.821 & 0.948 & 0.745 \\
& D-Y-V1    & 0.248 & 0.422 & 0.278 & 0.452 & 0.421 & 0.665 & 0.483 & 0.719 & 0.515 & 0.731 & 0.407 & 0.512 & 0.731 & 0.404 \\
& D-Y-V2    & 0.377 & 0.591 & 0.407 & 0.621 & 0.662 & 0.861 & 0.821 & 0.949 & 0.876 & 0.975 & 0.819 & 0.872 & 0.973 & 0.813 \\
& EN-DE-V1  & 0.302 & 0.492 & 0.332 & 0.522 & 0.514 & 0.710 & 0.555 & 0.744 & 0.580 & 0.759 & 0.484 & 0.585 & 0.769 & 0.486 \\
& EN-DE-V2  & 0.409 & 0.616 & 0.439 & 0.646 & 0.704 & 0.876 & 0.762 & 0.893 & 0.790 & 0.917 & 0.715 & 0.786 & 0.914 & 0.711 \\
& EN-FR-V1  & 0.229 & 0.401 & 0.259 & 0.431 & 0.378 & 0.596 & 0.405 & 0.638 & 0.480 & 0.692 & 0.374 & 0.479 & 0.691 & 0.373 \\
& EN-FR-V2  & 0.484 & 0.746 & 0.514 & 0.776 & 0.589 & 0.784 & 0.666 & 0.866 & 0.834 & 0.961 & 0.758 & 0.840 & 0.962 & 0.768 \\
\midrule
\multirow{5}{*}{DBP}
& FR-EN 15K & 0.075 & 0.245 & 0.075 & 0.245 & 0.139 & 0.261 & 0.343 & 0.572 & 0.454 & 0.638 & 0.356 & 0.464 & 0.652 & 0.363 \\
& JA-EN 15K & 0.095 & 0.294 & 0.095 & 0.294 & 0.147 & 0.283 & 0.332 & 0.543 & 0.442 & 0.607 & 0.353 & 0.450 & 0.618 & 0.358 \\
& ZH-EN 15K & 0.084 & 0.245 & 0.084 & 0.245 & 0.157 & 0.291 & 0.327 & 0.540 & 0.409 & 0.579 & 0.318 & 0.414 & 0.585 & 0.323 \\
& D-W-100K & 0.452 & 0.621 & 0.452 & 0.671 & 0.453 & 0.657 & 0.602 & 0.788 & 0.687 & 0.846 & 0.600 & 0.737 & 0.880 & 0.655 \\
& D-Y-100K & 0.310 & 0.488 & 0.410 & 0.588 & 0.573 & 0.771 & 0.718 & 0.867 & 0.818 & 0.941 & 0.748 & 0.823 & 0.944 & 0.755 \\
\bottomrule
\end{tabular}
}
\end{table*}

\section{Detailed Ablation Results on Structure Views}
\label{appx:structureviews}
Table~\ref{tab:decoder_feature_ablation_masking_synth} reports the detailed results of the structure-view ablation in our decoder.
In most cases, removing any one view leads to a performance drop, which again suggests that entity, neighborhood, relation, and anchor information are all useful for stable decoding.

\section{Additional Results on Gold vs. Hard-Negative Score Gaps}
\label{sec:appendix_gap_distributions}

\begin{figure*}[p]
\centering
\includegraphics[width=0.19\textwidth]{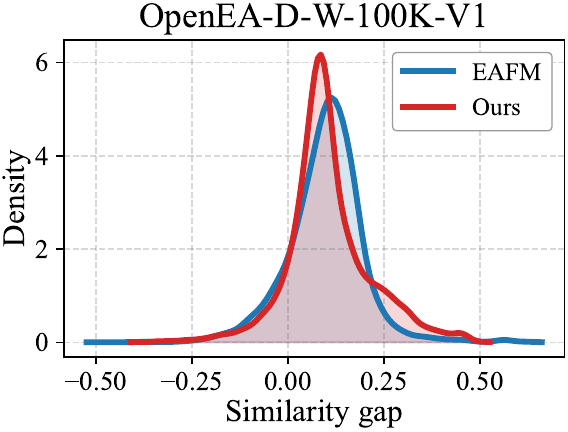}\hfill
\includegraphics[width=0.19\textwidth]{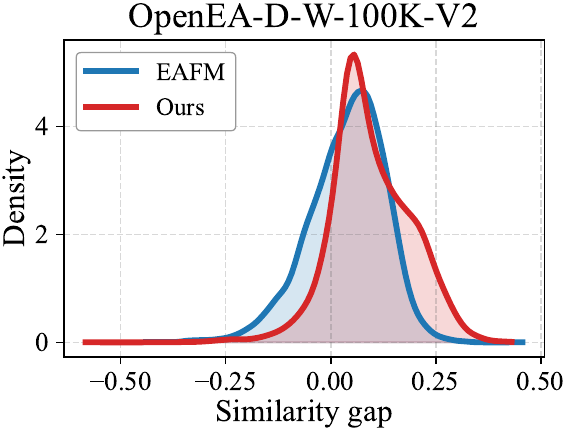}\hfill
\includegraphics[width=0.19\textwidth]{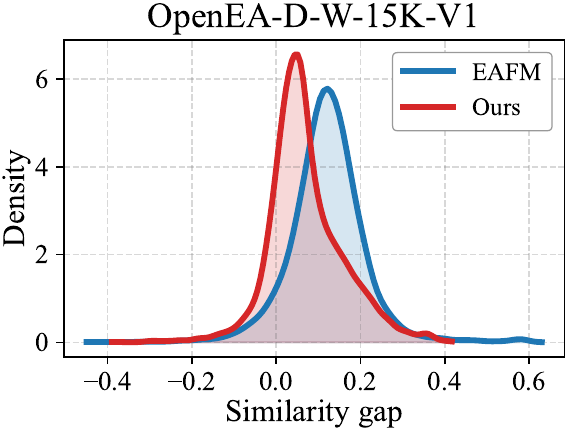}\hfill
\includegraphics[width=0.19\textwidth]{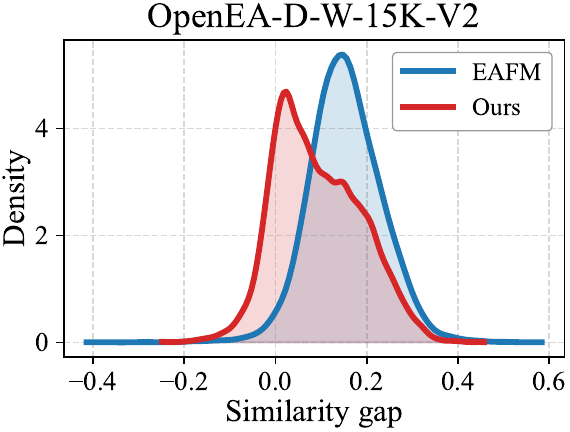}\hfill
\includegraphics[width=0.19\textwidth]{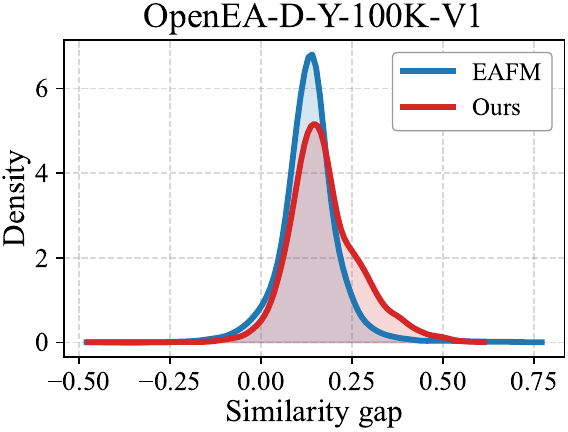}

\vspace{0.5em}
\includegraphics[width=0.19\textwidth]{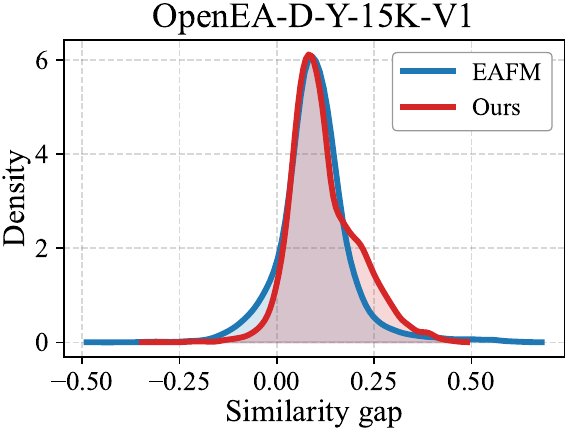}\hfill
\includegraphics[width=0.19\textwidth]{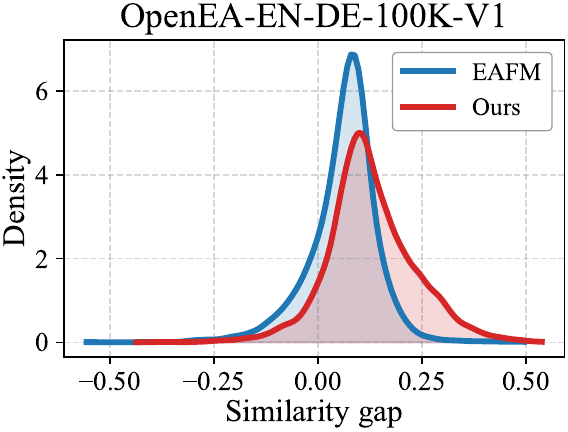}\hfill
\includegraphics[width=0.19\textwidth]{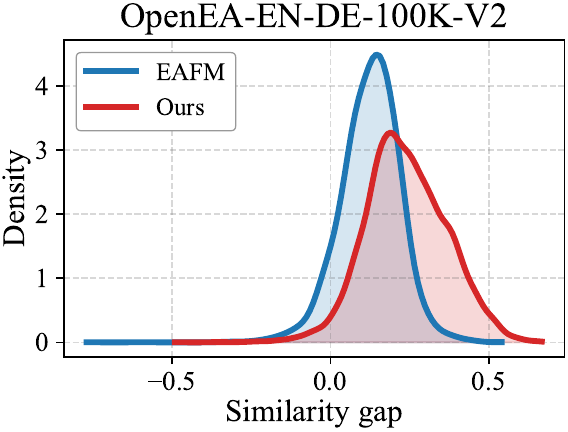}\hfill
\includegraphics[width=0.19\textwidth]{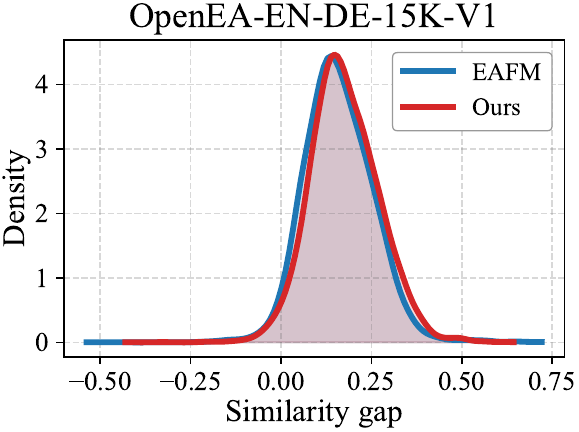}\hfill
\includegraphics[width=0.19\textwidth]{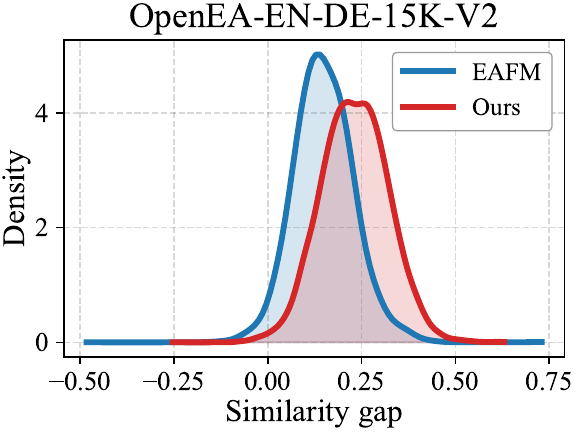}

\vspace{0.5em}
\includegraphics[width=0.19\textwidth]{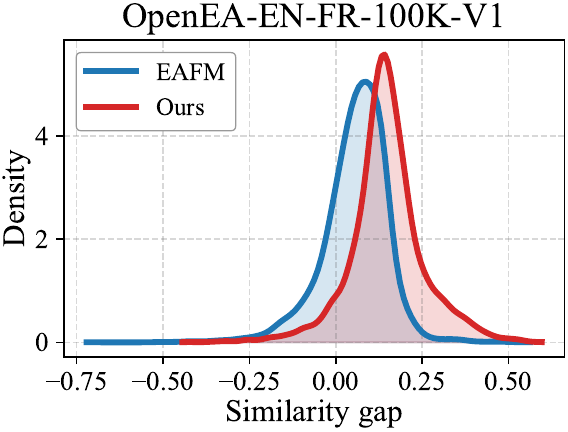}\hfill
\includegraphics[width=0.19\textwidth]{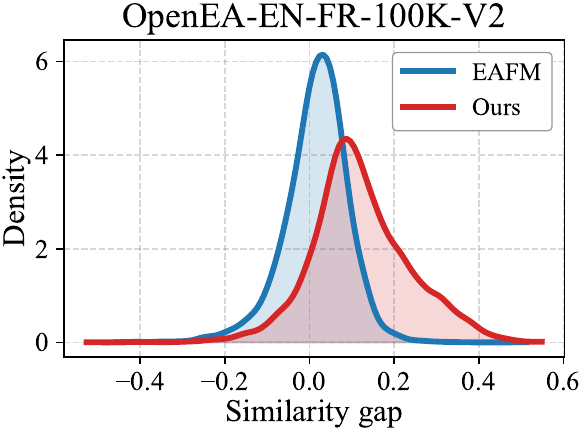}\hfill
\includegraphics[width=0.19\textwidth]{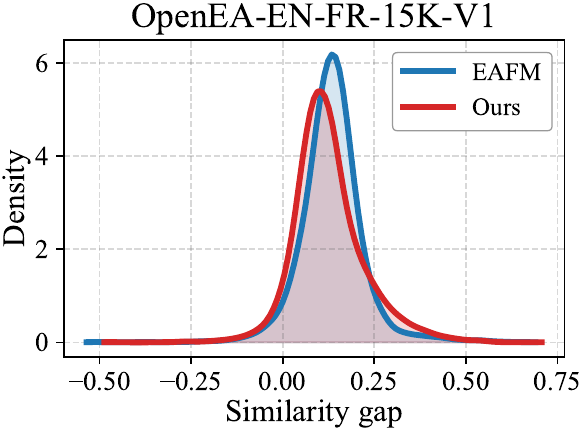}\hfill
\includegraphics[width=0.19\textwidth]{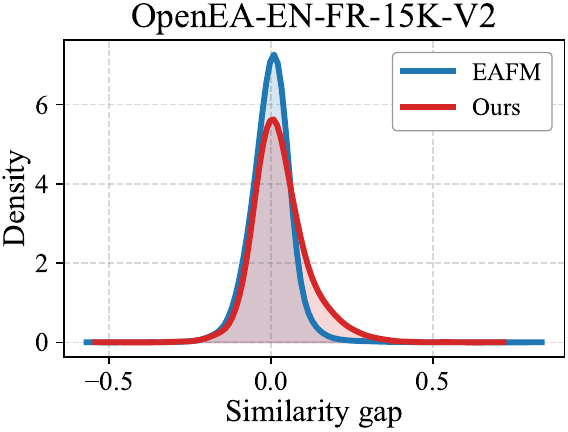}\hfill
\includegraphics[width=0.19\textwidth]{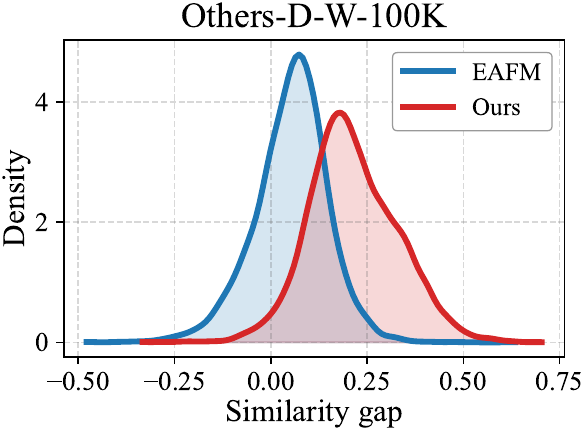}

\vspace{0.5em}
\includegraphics[width=0.19\textwidth]{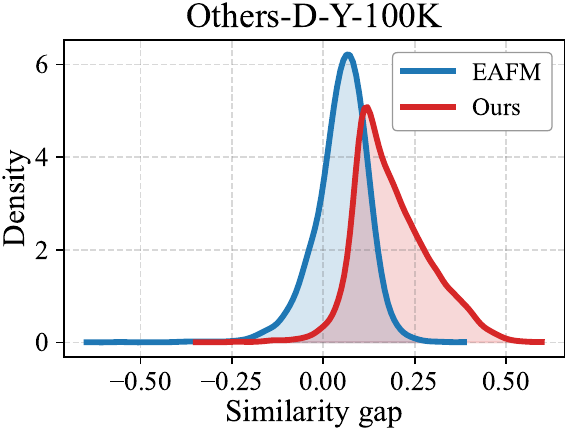}\hfill
\includegraphics[width=0.19\textwidth]{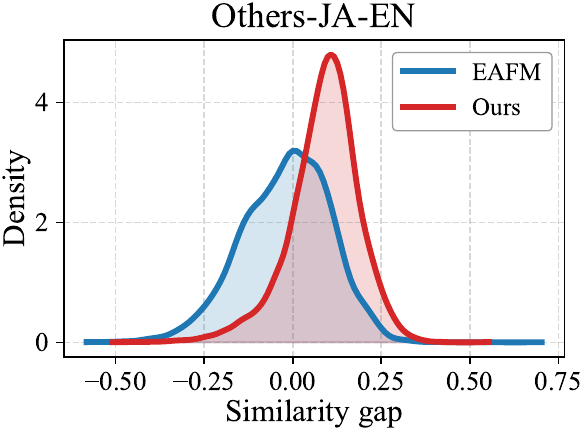}\hfill
\includegraphics[width=0.19\textwidth]{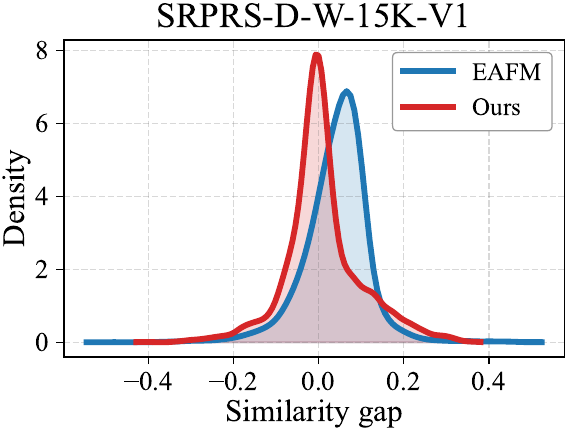}\hfill
\includegraphics[width=0.19\textwidth]{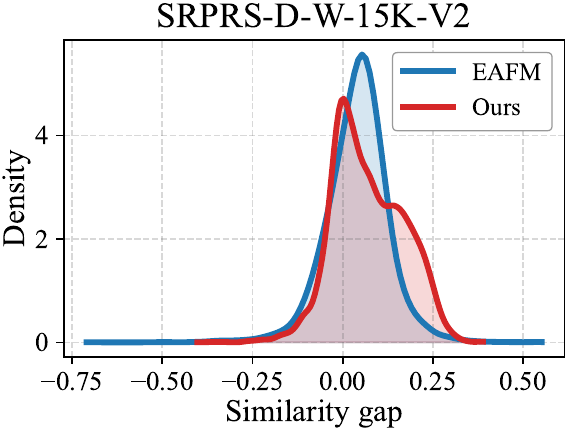}\hfill
\includegraphics[width=0.19\textwidth]{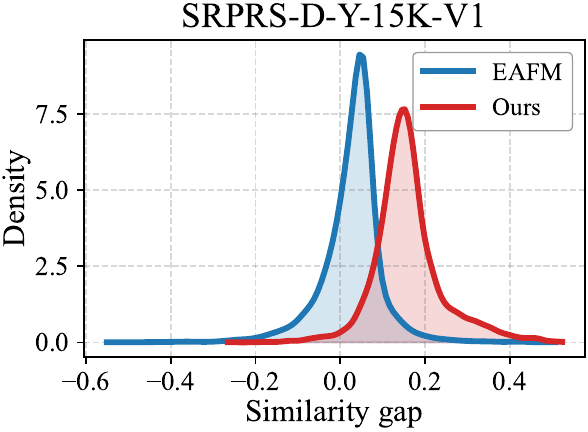}

\vspace{0.5em}
\includegraphics[width=0.19\textwidth]{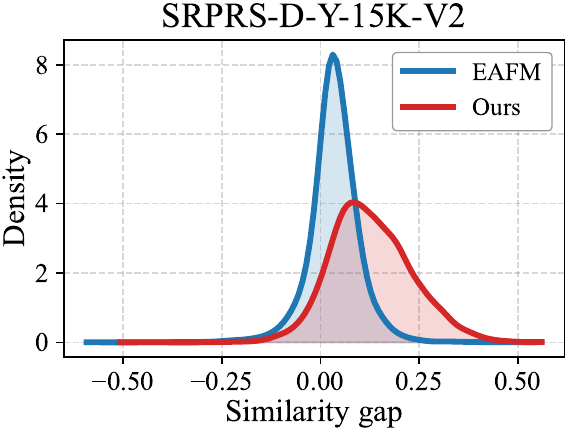}\hfill
\includegraphics[width=0.19\textwidth]{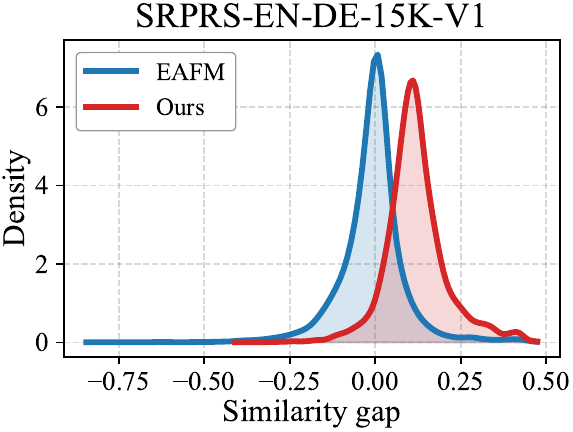}\hfill
\includegraphics[width=0.19\textwidth]{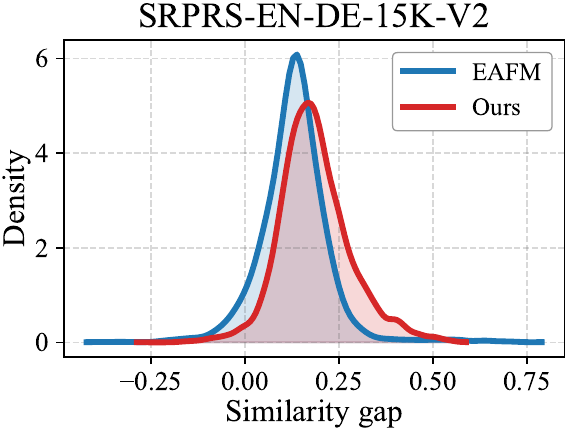}\hfill
\includegraphics[width=0.19\textwidth]{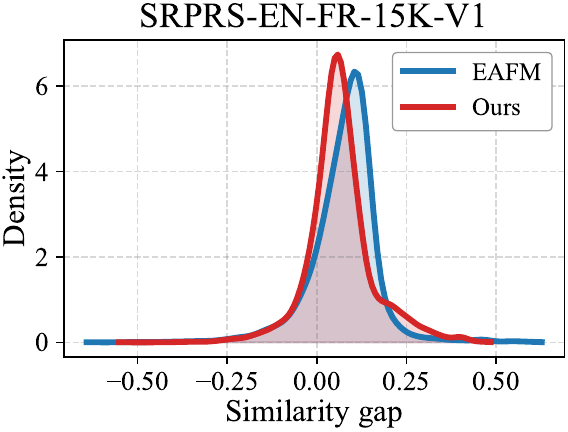}\hfill
\includegraphics[width=0.19\textwidth]{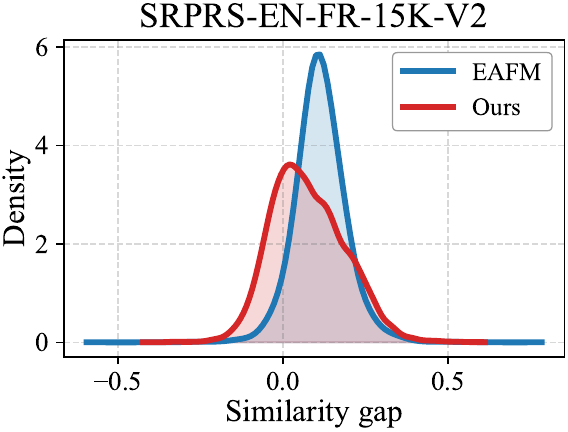}
\caption{Additional results on gold vs. hard-negative score gaps.}
\label{fig:gap_appendix}
\end{figure*}

Figure~\ref{fig:gap_appendix} reports the similarity-gap distributions for the remaining datasets not shown in the main text. 
The same interpretation applies as in Figure~\ref{fig:gap_distribution}: 
a larger positive gap indicates clearer separation between the gold target and the hardest negative candidate.

\section{Detailed Results on Efficiency}
\label{appx:efficiency}
Here we give more detailed efficiency results. Table~\ref{tab:per_dataset_time_mrr_aligned} reports per-dataset results, and Table~\ref{tab:efficiency_tradeoff_avg} reports the group-wise and overall averages. 
The results are consistent with Figure~\ref{fig:efficiency_tradeoff}: the encoder-only variant is the fastest, while the decoder brings extra gains at the cost of higher inference time when $k$ becomes larger.

\begin{table*}[t]
\centering
\caption{Per-dataset efficiency and effectiveness comparison. Inference time per query is measured directly. MRR for the EAFM baseline and the $k=16$ setting of \modelname\ is taken from Table~\ref{tab:detailed_ea_results}.}
\resizebox{\textwidth}{!}{\large
\begin{tabular}{llcccccccccccc}
\toprule
\multirow{2}{*}{\textbf{Group}} & \multirow{2}{*}{\textbf{Dataset}}
& \multicolumn{2}{c}{\textbf{EAFM}}
& \multicolumn{2}{c}{\textbf{\modelname}}
& \multicolumn{2}{c}{\textbf{\modelname}}
& \multicolumn{2}{c}{\textbf{\modelname}}
& \multicolumn{2}{c}{\textbf{\modelname}}
& \multicolumn{2}{c}{\textbf{\modelname}} \\
&
& \multicolumn{2}{c}{{pretrain}}
& \multicolumn{2}{c}{{w/o Decoder}}
& \multicolumn{2}{c}{\textbf{$k=8$}}
& \multicolumn{2}{c}{\textbf{$k=16$}}
& \multicolumn{2}{c}{\textbf{$k=32$}}
& \multicolumn{2}{c}{\textbf{$k=64$}} \\
\cmidrule(lr){3-4}\cmidrule(lr){5-6}\cmidrule(lr){7-8}\cmidrule(lr){9-10}\cmidrule(lr){11-12}\cmidrule(lr){13-14}
& & Time (s) & MRR & Time (s) & MRR & Time (s) & MRR & Time (s) & MRR & Time (s) & MRR & Time (s) & MRR \\
\midrule
\multirow{16}{*}{OpenEA}
& D-W-15K-V1  & 0.011 & 0.604 & 0.008 & 0.662 & 0.015 & 0.671 & 0.021 & 0.674 & 0.033 & 0.676 & 0.058 & 0.673 \\
& D-W-15K-V2  & 0.017 & 0.738 & 0.011 & 0.830 & 0.018 & 0.839 & 0.026 & 0.842 & 0.038 & 0.844 & 0.062 & 0.840 \\
& D-Y-15K-V1  & 0.008 & 0.636 & 0.006 & 0.691 & 0.012 & 0.700 & 0.019 & 0.704 & 0.031 & 0.702 & 0.054 & 0.706 \\
& D-Y-15K-V2  & 0.013 & 0.950 & 0.008 & 0.959 & 0.015 & 0.969 & 0.022 & 0.971 & 0.034 & 0.970 & 0.060 & 0.973 \\
& D-W-100K-V1 & 0.051 & 0.459 & 0.042 & 0.535 & 0.046 & 0.545 & 0.055 & 0.548 & 0.066 & 0.550 & 0.090 & 0.546 \\
& D-W-100K-V2 & 0.144 & 0.437 & 0.081 & 0.680 & 0.086 & 0.691 & 0.095 & 0.694 & 0.107 & 0.696 & 0.133 & 0.692 \\
& D-Y-100K-V1 & 0.075 & 0.694 & 0.050 & 0.729 & 0.054 & 0.737 & 0.063 & 0.740 & 0.073 & 0.742 & 0.098 & 0.738 \\
& D-Y-100K-V2 & 0.155 & 0.762 & 0.089 & 0.895 & 0.094 & 0.905 & 0.103 & 0.908 & 0.115 & 0.907 & 0.143 & 0.910 \\
& EN-DE-15K-V1 & 0.011 & 0.737 & 0.007 & 0.775 & 0.014 & 0.783 & 0.021 & 0.786 & 0.034 & 0.788 & 0.058 & 0.785 \\
& EN-DE-15K-V2 & 0.019 & 0.765 & 0.011 & 0.874 & 0.018 & 0.882 & 0.025 & 0.886 & 0.038 & 0.884 & 0.065 & 0.889 \\
& EN-FR-15K-V1 & 0.011 & 0.533 & 0.008 & 0.607 & 0.014 & 0.617 & 0.021 & 0.620 & 0.033 & 0.622 & 0.056 & 0.618 \\
& EN-FR-15K-V2 & 0.018 & 0.460 & 0.011 & 0.709 & 0.018 & 0.719 & 0.026 & 0.723 & 0.038 & 0.721 & 0.060 & 0.725 \\
& EN-DE-100K-V1 & 0.074 & 0.520 & 0.050 & 0.561 & 0.054 & 0.571 & 0.063 & 0.573 & 0.073 & 0.575 & 0.098 & 0.571 \\
& EN-DE-100K-V2 & 0.136 & 0.594 & 0.079 & 0.734 & 0.084 & 0.743 & 0.093 & 0.746 & 0.105 & 0.748 & 0.131 & 0.744 \\
& EN-FR-100K-V1 & 0.065 & 0.364 & 0.046 & 0.425 & 0.050 & 0.436 & 0.059 & 0.439 & 0.070 & 0.437 & 0.095 & 0.441 \\
& EN-FR-100K-V2 & 0.141 & 0.378 & 0.079 & 0.573 & 0.083 & 0.585 & 0.093 & 0.588 & 0.104 & 0.589 & 0.129 & 0.586 \\
\midrule
\multirow{8}{*}{SRPRS}
& D-W-V1   & 0.012 & 0.340 & 0.008 & 0.503 & 0.015 & 0.513 & 0.022 & 0.514 & 0.035 & 0.516 & 0.057 & 0.512 \\
& D-W-V2   & 0.018 & 0.633 & 0.011 & 0.807 & 0.019 & 0.816 & 0.026 & 0.819 & 0.039 & 0.821 & 0.060 & 0.817 \\
& D-Y-V1   & 0.010 & 0.421 & 0.007 & 0.502 & 0.014 & 0.513 & 0.021 & 0.515 & 0.034 & 0.513 & 0.056 & 0.518 \\
& D-Y-V2   & 0.019 & 0.662 & 0.011 & 0.862 & 0.018 & 0.873 & 0.026 & 0.876 & 0.040 & 0.878 & 0.062 & 0.874 \\
& EN-DE-V1 & 0.011 & 0.514 & 0.007 & 0.568 & 0.014 & 0.578 & 0.021 & 0.580 & 0.034 & 0.582 & 0.055 & 0.579 \\
& EN-DE-V2 & 0.014 & 0.704 & 0.009 & 0.777 & 0.016 & 0.787 & 0.023 & 0.790 & 0.036 & 0.792 & 0.057 & 0.788 \\
& EN-FR-V1 & 0.011 & 0.378 & 0.008 & 0.469 & 0.015 & 0.478 & 0.021 & 0.480 & 0.033 & 0.482 & 0.057 & 0.478 \\
& EN-FR-V2 & 0.017 & 0.589 & 0.011 & 0.821 & 0.018 & 0.831 & 0.025 & 0.834 & 0.038 & 0.836 & 0.057 & 0.832 \\
\midrule
\multirow{5}{*}{DBP}
& FR-EN 15K & 0.088 & 0.139 & 0.067 & 0.443 & 0.073 & 0.452 & 0.082 & 0.454 & 0.091 & 0.456 & 0.114 & 0.452 \\
& JA-EN 15K & 0.082 & 0.147 & 0.065 & 0.430 & 0.070 & 0.439 & 0.078 & 0.442 & 0.087 & 0.444 & 0.110 & 0.440 \\
& ZH-EN 15K & 0.097 & 0.157 & 0.080 & 0.396 & 0.085 & 0.405 & 0.093 & 0.409 & 0.103 & 0.407 & 0.131 & 0.410 \\
& D-W-100K  & 0.102 & 0.453 & 0.066 & 0.673 & 0.071 & 0.684 & 0.080 & 0.687 & 0.090 & 0.689 & 0.113 & 0.685 \\
& D-Y-100K  & 0.101 & 0.573 & 0.064 & 0.803 & 0.069 & 0.814 & 0.079 & 0.818 & 0.088 & 0.816 & 0.110 & 0.820 \\
\bottomrule
\end{tabular}
}
\label{tab:per_dataset_time_mrr_aligned}
\end{table*}
\begin{table*}[t]
\centering
\caption{Average efficiency-effectiveness comparison under different inference settings. We report average MRR and average inference time (s) per query across dataset groups.}
\resizebox{0.85\linewidth}{!}{
\begin{tabular}{lcccccccc}
\toprule
\multirow{2}{*}{\textbf{Method}}
& \multicolumn{2}{c}{\textbf{OpenEA}}
& \multicolumn{2}{c}{\textbf{SRPRS}}
& \multicolumn{2}{c}{\textbf{DBP}}
& \multicolumn{2}{c}{\textbf{Average}} \\
\cmidrule(lr){2-3} \cmidrule(lr){4-5} \cmidrule(lr){6-7} \cmidrule(lr){8-9}
& Time (s) & MRR & Time (s) & MRR & Time (s) & MRR & Time (s) & MRR \\
\midrule
EAFM
& 0.059 & 0.602
& 0.014 & 0.530
& 0.094 & 0.294
& 0.053 & 0.529 \\
\modelname\ w/o Decoder
& {0.037} & 0.702
& {0.009} & 0.664
& {0.068} & 0.549
& {0.034} & 0.665 \\
\modelname\ ($k=8$)
& 0.042 & 0.712
& 0.016 & 0.674
& 0.074 & 0.559
& 0.040 & 0.675 \\
\modelname\ ($k=16$)
& 0.050 & 0.715
& 0.023 & 0.676
& 0.082 & 0.562
& 0.048 & 0.678 \\
\modelname\ ($k=32$)
& 0.062 & {0.716}
& 0.036 & {0.678}
& 0.092 & {0.562}
& 0.060 & {0.679} \\
\modelname\ ($k=64$)
& 0.087 & 0.715
& 0.058 & 0.675
& 0.116 & 0.561
& 0.084 & 0.677 \\
\bottomrule
\end{tabular}
}
\label{tab:efficiency_tradeoff_avg}
\end{table*}

\section{More Results of MRR w.r.t. Weight $\beta$}\label{appx:beta}
Table~\ref{tab:beta_sensitivity} reports more detailed MRR results under different values of $\beta$. 
The overall trend is stable, and a moderate value usually gives the best or near-best performance. This is consistent with the analysis in the main text.

\begin{table*}[t]
\centering
\caption{Sensitivity analysis of the structural calibration weight $\beta$ using representative test datasets.}
\resizebox{0.8\linewidth}{!}{
\begin{tabular}{lcccccc}
\toprule
\multirow{2}{*}{\textbf{$\beta$}}
& \multicolumn{5}{c}{\textbf{MRR}} 
& \multirow{2}{*}{\textbf{Average}} \\
\cmidrule(lr){2-6}
& \textbf{D-W-15K-V1}
& \textbf{D-W-15K-V2}
& \textbf{D-W-100K-V1}
& \textbf{EN-DE-15K-V1}
& \textbf{FR-EN}
&  \\
\midrule
0.1
& 0.671 & 0.827 & 0.554 & 0.785 & 0.343 & 0.636 \\
0.3
& 0.665 & {0.849} & {0.563} & 0.780 & {0.430} & {0.657} \\
0.5
& 0.665 & {0.845} & 0.539 & {0.787} & 0.380 & {0.643} \\
1.0
& {0.674} & 0.833 & 0.536 & 0.770 & {0.407} & 0.644 \\
\bottomrule
\end{tabular}
}
\label{tab:beta_sensitivity}
\end{table*}

\begin{table*}[t]
\centering
\caption{Detailed ablation results on structure views used in our decoder.}
\resizebox{0.9\textwidth}{!}{
\begin{tabular}{llcccccccc}
\toprule
\multirow{2}{*}{\textbf{Group}} & \multirow{2}{*}{\textbf{Dataset}}
& \multicolumn{2}{c}{\textbf{w/o Entity}}
& \multicolumn{2}{c}{\textbf{w/o Neighborhood}}
& \multicolumn{2}{c}{\textbf{w/o Relation}}
& \multicolumn{2}{c}{\textbf{w/o Anchor}} \\
\cmidrule(lr){3-4}\cmidrule(lr){5-6}\cmidrule(lr){7-8}\cmidrule(lr){9-10}
& & MRR & H@1 & MRR & H@1 & MRR & H@1 & MRR & H@1 \\
\midrule
\multirow{17}{*}{OpenEA}
& D-W-100K-V1 & 0.555 & 0.447 & 0.532 & 0.459 & 0.553 & 0.461 & 0.549 & 0.457 \\
& D-W-100K-V2 & 0.707 & 0.619 & 0.696 & 0.601 & 0.691 & 0.593 & 0.701 & 0.611 \\
& D-W-15K-V1 & 0.658 & 0.565 & 0.656 & 0.568 & 0.666 & 0.581 & 0.666 & 0.573 \\
& D-W-15K-V2 & 0.818 & 0.767 & 0.814 & 0.761 & 0.812 & 0.761 & 0.811 & 0.755 \\
& D-Y-100K-V1 & 0.723 & 0.657 & 0.725 & 0.654 & 0.724 & 0.658 & 0.714 & 0.635 \\
& D-Y-100K-V2 & 0.893 & 0.831 & 0.878 & 0.850 & 0.874 & 0.827 & 0.874 & 0.842 \\
& D-Y-15K-V1 & 0.676 & 0.619 & 0.682 & 0.611 & 0.681 & 0.604 & 0.674 & 0.625 \\
& D-Y-15K-V2 & 0.967 & 0.942 & 0.958 & 0.933 & 0.966 & 0.952 & 0.949 & 0.945 \\
& EN-DE-100K-V1 & 0.564 & 0.464 & 0.579 & 0.478 & 0.575 & 0.489 & 0.562 & 0.469 \\
& EN-DE-100K-V2 & 0.716 & 0.624 & 0.701 & 0.635 & 0.713 & 0.634 & 0.705 & 0.644 \\
& EN-DE-15K-V1 & 0.760 & 0.684 & 0.766 & 0.691 & 0.768 & 0.687 & 0.772 & 0.681 \\
& EN-DE-15K-V2 & 0.851 & 0.783 & 0.861 & 0.786 & 0.857 & 0.803 & 0.852 & 0.806 \\
& EN-FR-100K-V1 & 0.422 & 0.338 & 0.414 & 0.316 & 0.419 & 0.327 & 0.430 & 0.321 \\
& EN-FR-100K-V2 & 0.607 & 0.505 & 0.585 & 0.492 & 0.586 & 0.506 & 0.592 & 0.495 \\
& EN-FR-15K-V1 & 0.583 & 0.468 & 0.583 & 0.465 & 0.583 & 0.478 & 0.592 & 0.470 \\
& EN-FR-15K-V2 & 0.688 & 0.572 & 0.692 & 0.569 & 0.693 & 0.586 & 0.697 & 0.582 \\
& Average & 0.699 & 0.618 & 0.695 & 0.617 & 0.698 & 0.622 & 0.696 & 0.619 \\
\midrule
\multirow{9}{*}{SRPRS}
& D-W-15K-V1 & 0.494 & 0.396 & 0.486 & 0.389 & 0.494 & 0.403 & 0.496 & 0.402 \\
& D-W-15K-V2 & 0.813 & 0.725 & 0.815 & 0.724 & 0.807 & 0.730 & 0.795 & 0.738 \\
& D-Y-15K-V1 & 0.488 & 0.390 & 0.487 & 0.378 & 0.493 & 0.387 & 0.485 & 0.391 \\
& D-Y-15K-V2 & 0.873 & 0.816 & 0.862 & 0.816 & 0.868 & 0.808 & 0.856 & 0.806 \\
& EN-DE-15K-V1 & 0.564 & 0.459 & 0.572 & 0.452 & 0.567 & 0.460 & 0.567 & 0.461 \\
& EN-DE-15K-V2 & 0.789 & 0.703 & 0.780 & 0.698 & 0.788 & 0.692 & 0.772 & 0.703 \\
& EN-FR-15K-V1 & 0.459 & 0.346 & 0.457 & 0.345 & 0.451 & 0.362 & 0.452 & 0.351 \\
& EN-FR-15K-V2 & 0.810 & 0.720 & 0.798 & 0.733 & 0.799 & 0.736 & 0.808 & 0.718 \\
& Average & 0.661 & 0.569 & 0.657 & 0.567 & 0.658 & 0.572 & 0.654 & 0.571 \\
\midrule
\multirow{6}{*}{DBP}
& D-W-100K & 0.739 & 0.658 & 0.740 & 0.655 & 0.731 & 0.654 & 0.739 & 0.642 \\
& D-Y-100K & 0.833 & 0.766 & 0.831 & 0.766 & 0.834 & 0.780 & 0.840 & 0.769 \\
& FR-EN & 0.156 & 0.092 & 0.138 & 0.109 & 0.153 & 0.105 & 0.142 & 0.110 \\
& JA-EN & 0.141 & 0.098 & 0.142 & 0.114 & 0.151 & 0.114 & 0.148 & 0.113 \\
& ZH-EN & 0.106 & 0.073 & 0.091 & 0.072 & 0.110 & 0.078 & 0.105 & 0.066 \\
& Average & 0.395 & 0.337 & 0.388 & 0.343 & 0.396 & 0.346 & 0.395 & 0.340 \\
\midrule
\multirow{1}{*}{Overall}
& Average & 0.636 & 0.556 & 0.632 & 0.556 & 0.635 & 0.561 & 0.633 & 0.558 \\
\bottomrule
\end{tabular}
}
\label{tab:decoder_feature_ablation_masking_synth}
\end{table*}

\section{Limitations}
Our method still has several limitations. First, the evaluation is based on existing EA benchmarks, so the conclusions may still depend on the benchmark construction and data quality. Second, the method relies on training anchors to support cross-KG interaction and structural calibration. When anchors are very sparse or noisy, the benefit may become smaller. Third, although the decoder is lightweight, its inference cost still grows with the candidate size $k$.

\section{Broader Impacts}
This work mainly improves transferable EA on public KG benchmarks. A possible positive impact is that better EA can support knowledge fusion and improve the quality of downstream knowledge-driven systems. A possible risk is that stronger alignment tools may also be used to link entities across data sources in settings where data integration should be carefully controlled. We therefore view this work as a general technical contribution, and responsible use should depend on the application context and data governance requirements.

\end{document}